\documentclass[sigconf,nonacm]{aamas} 

\usepackage{balance} % for balancing columns on the final page
\usepackage{bm}

\usepackage{algorithm}
\usepackage{algorithmic} 
\usepackage{subfigure}
\usepackage{graphicx}
\usepackage{epstopdf}
\usepackage{multirow}

\setcopyright{ifaamas}
\acmConference[AAMAS '22]{Proc.\@ of the 21st International Conference
on Autonomous Agents and Multiagent Systems (AAMAS 2022)}{May 9--13, 2022}
{Auckland, New Zealand}{P.~Faliszewski, V.~Mascardi, C.~Pelachaud,
M.E.~Taylor (eds.)}
\copyrightyear{2022}
\acmYear{2022}
\acmDOI{}
\acmPrice{}
\acmISBN{}

\acmSubmissionID{???}

\title[AAMAS-2022 Formatting Instructions]{PooL: Pheromone-inspired Communication Framework for Large Scale Multi-Agent Reinforcement Learning}

\author{Zixuan Cao}
\affiliation{
  \institution{Peking University}
  \city{Beijing}
  \country{China}}
\email{caozixuan.percy@stu.pku.edu.cn}

\author{Mengzhi Shi}
\affiliation{
  \institution{Peking University}
  \city{Beijing}
  \country{China}}
\email{shimengzhi@pku.edu.cn}

\author{Zhanbo Zhao}
\affiliation{
  \institution{Peking University}
  \city{Beijing}
  \country{China}}
\email{zhaozb1997@pku.edu.cn}

\author{Xiujun Ma}
\affiliation{
  \institution{Peking University}
  \city{Beijing}
  \country{China}}
\email{maxj@cis.pku.edu.cn}

\begin{abstract}

Being difficult to scale poses great problems in multi-agent coordination. Multi-agent Reinforcement Learning (MARL) algorithms applied in small-scale multi-agent systems are hard to extend to large-scale ones because the latter is far more dynamic and the number of interactions increases exponentially with the growing number of agents. Some swarm intelligence algorithms simulate the release and utilization mechanism of pheromones to control large-scale agent coordination. Inspired by such algorithms, \textbf{PooL}, an \textbf{p}her\textbf{o}m\textbf{o}ne-based indirect communication framework applied to large scale multi-agent reinforcement \textbf{l}earning is proposed in order to solve the large-scale multi-agent coordination problem. Pheromones released by agents of PooL are defined as outputs of most reinforcement learning algorithms, which reflect agents' views of the current environment. The pheromone update mechanism can efficiently organize the information of all agents and simplify the complex interactions among agents into low-dimensional representations. Pheromones perceived by agents can be regarded as a summary of the views of nearby agents which can better reflect the real situation of the environment. Q-Learning is taken as our base model to implement PooL and PooL is evaluated in various large-scale cooperative environments. Experiments show agents can capture effective information through PooL and achieve higher rewards than other state-of-arts methods with lower communication costs.
\end{abstract}


\newcommand{\BibTeX}{\rm B\kern-.05em{\sc i\kern-.025em b}\kern-.08em\TeX}

\begin{document}

%%% The following commands remove the headers in your paper. For final 
%%% papers, these will be inserted during the pagination process.

\pagestyle{fancy}
\fancyhead{}

%%% The next command prints the information defined in the preamble.

\maketitle 

%%%%%%%%%%%%%%%%%%%%%%%%%%%%%%%%%%%%%%%%%%%%%%%%%%%%%%%%%%%%%%%%%%%%%%%%

\section{Introduction}

MARL focuses on developing algorithms to optimize behaviors of different agents which share a common environment~\cite{bucsoniu2010multi}. Agents' behaviors in MARL settings are usually modeled as Partially Observable Markov Decision Process (POMDP), in which agents only have a partial observation of the global state. MARL faces two major problems. First, partial observations from different agents as input features cause a curse of dimensionality. Second, it is difficult to consider all possible interactions among agents because the number of possible interactions grows exponentially. With the increase of agents' scale, these two problems will become more and more serious. In MARL, there are various methods to solve the two problems. These methods can be roughly divided into two categories, Centralized Training Decentralized Execution (CTDE)~\cite{oliehoek2008optimal} and multi-agent communication. In the CTDE framework, agents can access global information while training, and agents select actions based on partial observations while execution. But CTDE algorithms are still unable to deal with large-scale agents for the reason that the dimension of global information is still too high to handle even at the training step. Multi-agent communication is another promising approach to coordinate the behaviors of large-scale agents. Agents can choose actions by their partial observations and information received from other agents. But designing an effective communication mechanism is not a trivial task. For an agent, the message representation, message sender selection, and message utilization all need to be carefully designed. In addition, agents' communication is limited by bandwidth in reality.  This bandwidth limitation also affects the scale of multi-agent communication.

Swarm Intelligence refers to solving problems by the interactions of simple information-processing units~\cite{kennedy2006swarm}. One of the key concepts of Swarm Intelligence is Stigmergy which is proposed in the 1950s by Pierre-Paul Grassé~\cite{heylighen2015stigmergy}. Summarizing from behaviors of insects, Stigmergy refers to a mechanism of indirect coordination between agents by the trace left in the environment. A representative algorithm inspired by Stigmergy is Ant Colony Optimization (ACO) algorithm. It takes inspiration from the foraging behavior of some ant species and uses a mechanism similar to pheromones for solving optimization problems~\cite{dorigo2006ant}. ACO is widely used in various path-finding problems in the real world~\cite{konatowski2018ant, dai2019mobile}. The effectiveness of such ACO algorithms depends on the exploration of a large number of units (e.g. robots) and pheromone-based indirect communication among them. However, in ACO's settings, units' behaviors are confined by predefined rules, which is unsuitable in MARL environments.

In this paper, inspired by the pheromone mechanism introduced in ACO, an indirect communication framework for MARL is developed by introducing pheromones into Deep Reinforcement Learning (DRL). The output of reinforcement learning algorithms is usually the scoring of actions or the probabilities of selecting a certain action. These values can be used not only to select actions but also to reflect the agents' views of the environment. These values output by different agents can be organized by a pheromone update mechanism similar to ACO. Therefore, the pheromone information perceived by agents combines the views and knowledge of other agents around them, and can better reflect the real situation of the current agent. 
\begin{figure}[h]
  \centering
  \includegraphics[width=1.0\linewidth]{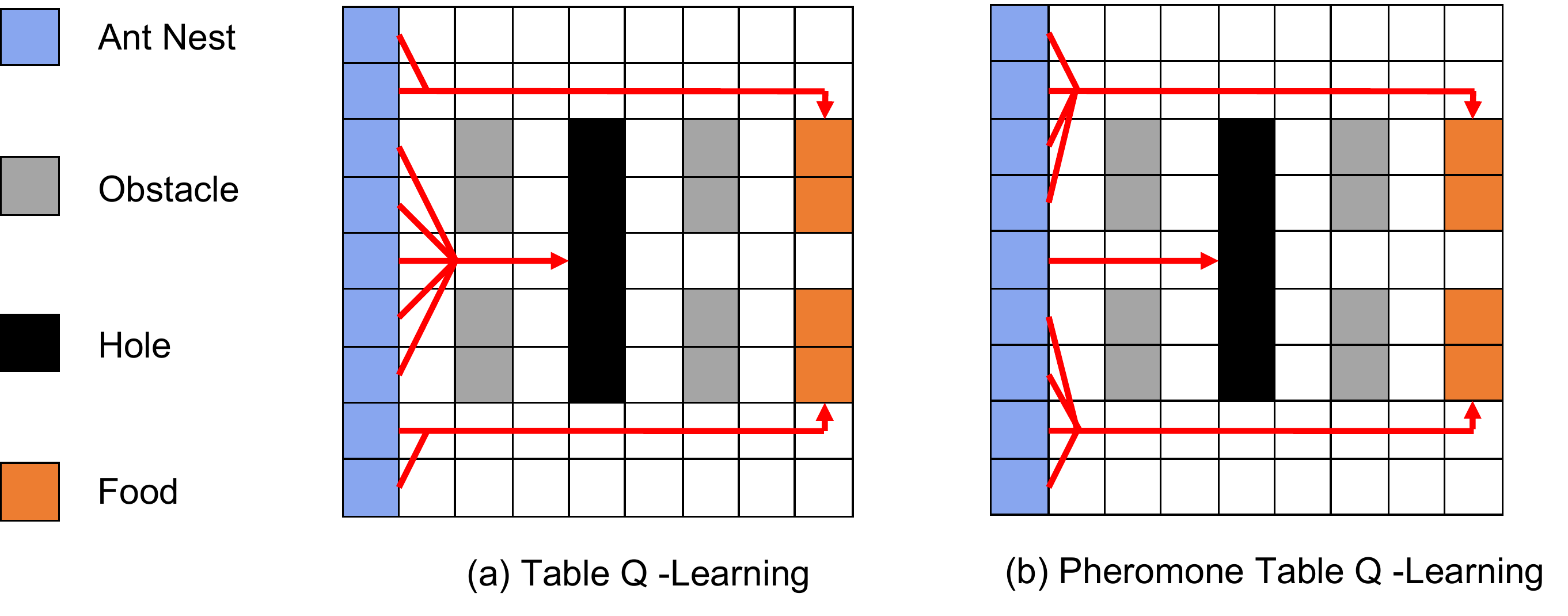}
  \caption{Intuition behind PooL (a) Most agents trained by Table Q-Learning without any communication fall into holes. (b) More than half agents find food with the guidance of pheromones by pheromone based Q-Learning.}
  \label{fig:intro_motivation}
  \Description{Intuition behind PooL (a) Most agents trained by Table Q-Learning without any communication fall into holes. (b) More than half agents find food with the guidance of pheromones by pheromone based Table Q-Learning.}
\end{figure}

The intuition behind our framework is shown in Figure \ref{fig:intro_motivation}. In a simple grid maze environment, there are ant nests (blue grids), obstacles (grey grids), holes (black grids), and food (orange grids). Agents (i.e. ants) get negative rewards at every time step. Agents will exit from the environment if they fall into holes or find food. Agents who find food get extra positive rewards. If agents take actions dependently, they are very easy to fall into local optimization (fall into holes). But if agents communicate through pheromones, those agents who find the optimal solution will transfer the information of the optimal solution to other agents. As a result, more agents find the optimal solution with the guidance of pheromones. More details about this motivation example will be discussed in Section 6.
% we need to rewrite this picture, the red line means the way forward

Following the intuition above, the pheromone mechanism is extended to the field of DRL. PooL simulates the process of ant colonies releasing pheromone, feeling the pheromone and the surrounding environment, and making decisions. A pheromone receptor is designed to convert pheromone information to a fixed dimension feature. For each agent, the original reinforcement learning component receives information from the real world and the pheromone receptor to choose actions and release pheromones. In this way, PooL realizes indirect agent communication and distributed training and execution in MARL. We combine PooL with Deep Q-Learning and evaluate PooL in a variety of settings of MAgent~\cite{zheng2018magent} which is further capsuled by PettingZoo~\cite{terry2020pettingzoo}. Experiments show that PooL can achieve effective communication of local information and get higher rewards.

The contributions of our paper can be summarized as follows:
\begin{enumerate}
\item We propose a pheromone-based indirect communication framework that combines MARL with swarm intelligence algorithms.  
\item PooL realizes more effective large-scale multi-agent coordination and achieves better performance than the existing methods when there are hundreds of agents in the environment.
\item PooL can be combined with any reinforcement learning algorithms, and the additional cost is very small, which makes it easier to apply to real-world scenes. 
\end{enumerate}

Our paper is organized as follows. Section 2 introduces work that is closely related to our proposed framework. Section 3 introduces the background knowledge related to our problem to be solved. In Section 4, we give a detailed description of PooL. In section 5, we show the performance of PooL in a variety of settings. Section 6 summarizes our paper. 
%%%%%%%%%%%%%%%%%%%%%%%%%%%%%%%%%%%%%%%%%%%%%%%%%%%%%%%%%%%%%%%%%%%%%%%%

\section{Related Work}
Our paper is closely related to two branches of study: Multi-agent Reinforcement Learning (MARL) and control algorithms that are based on Swarm Intelligence.

\subsection{Multi-agent Reinforcement Learning}

DRL methods like Deep Q-Learning (DQN)~\cite{mnih2013playing} promote the application of reinforcement learning in many fields such as Atari Games~\cite{mnih2015human} and Go~\cite{silver2016mastering}. But most existing methods of DRL focus on the control of a single agent. As in real life, many scenes contain multi-agent interactions and cooperation. MARL attracts more and more attention. \cite{lowe2020multi-agent,wei2018multiagent,foerster2018counterfactual} are classic MARL algorithms following the framework of CTDE~\cite{oliehoek2008optimal}. These algorithms apply modified single-agent reinforcement learning algorithms such as DDPG~\cite{lillicrap2016continuous} and Soft Q-Learning~\cite{haarnoja2017reinforcement} in MARL settings. However, such methods still suffer from the curse of dimensionality because the method still needs to handle all agents' features while training. Communication is an effective way to solve such problems. CommNet~\cite{sukhbaatar2016learning} designs a channel for all agents to communicate and extract information from all agents by average pooling. In order to better handle the situation with more agents,~\cite{zhao2019learning} uses an information filtering network to select useful information. VAIN~\cite{hoshen2018vain}, ATOC~\cite{jiang2018learning} and TarMAC~\cite{das2019tarmac} use attention mechanism to select communication targets and information more reasonably. But these methods still assume a centralized controller that can access all communication needed information while training. DGN~\cite{jiang2020graph} models agents' interactions with graph structures by Graph Neural Networks (GNN)~\cite{zhou2020graph}. In DGN, agents' communication is achieved by the convolution of graph nodes(e.g. agents nearby). Graph convolution is able to express more complex agent interactions. But its performance and convergence speed are still confined by the number of agents.

In order to make algorithms scale to many agents, MFQ \cite{luo*2018mean} accepts the average action value of nearby agents as extra network input. \cite{zhou2019factorized} proposes to approximate Q-function with factorized pairwise interactions. \cite{wang2020few} provides another idea that trains agents from few to more with transfer mechanisms.
\subsection{Control Algorithms Based on Swarm Intelligence}
Swarm Intelligence is widely used in group control problems. \cite{di1998antnet, maru2016qoe} discusses the use of swarm intelligence in adaptive learning of routing tables and network-failure-detection. \cite{bourenane2007reinforcement,xu2019brain} utilize pheromones to achieve multi-agent coordination. However, Swarm Intelligence algorithms like Bird Swarm Algorithm and Ant Colony Optimization are rule-based, as a result, they can not learn from environments. Phe-Q \cite{monekosso2001phe} first introduces the idea of pheromone mechanisms to reinforcement learning. \cite{matta2019q, meerza2019q} follow similar ideas and improve the convergence speed of Q-Learning in different environments. \cite{xu2021stigmergic} proposes to release the pheromone according to the distance between the agent and the target, and apply the pheromone information to Deep Reinforcement Learning.

Learning-based methods are more and more popular with the success of deep learning in many fields. Most methods mentioned above can not take advantage of deep neural networks. DRL is applied in \cite{xu2021stigmergic}, but the way \cite{xu2021stigmergic} obtains pheromones is artificially designed based on the specific environment. It will be more appropriate if pheromones are generated based on learning algorithms.

\section{Background}

\subsection{Problem Formulation}

The problem for multi-agent coordination in this paper can be considered as a Dec-POMDP game~\cite{hansen2004dynamic}. It can be well defined by a tuple $M=<I, S,\left\{A_{i}\right\}, P, \left\{R_{i}\right\},\left\{\Omega_{i}\right\}, O, T, \gamma>$, where $I$ is a set of $n$ agents, $S$ is the state space, $A_i$ is the action space for agent $i$, $\bm{\Omega_{i}}$ is the observation space for agent $i$, $T$ is the time horizon for the game and $\gamma$ is the discount factor for individual rewards ${R_i}$. When agents took actions $a$ in state $s$, the probability of the environment transitioning to state $s'$ is $P(s'|s,a)$ and the probability of agents seeing observations $o$ is $O(o|s,a)$.

For agents in the problem above, their goal is to maximize their own expected return $G_{i}=\sum_{t=0}^{T} \gamma^{t} R_{i}^{t}$. The achievement of this goal requires the cooperation of agents belonging to the same team.

\subsection{Table Q-Learning and Deep Q-Learning}

Q-Learning is a popular method in single agent reinforcement learning. For agent $i$, Q-Learning uses an action-value function for policy $\pi_i$ as $Q^{\pi}_{i}(\Omega_{i},a)=\mathbb{E}\left[R_{i} \mid \Omega^{t}_{i}=\Omega, a^{t}_{i}=a\right]$. Table Q-Learning maintains a Q-table and update Q-values by the following equation:
\begin{equation}
\label{eq:table_q}
     Q^{\pi}_{i}(\Omega_{i},a)=Q^{\pi}_{i}(\Omega_{i},a)+\alpha(R_{i}+\gamma \max_{a^{\prime}}Q^{\pi}_{i}(\Omega_{i}^{\prime},a^{\prime})-Q^{\pi}_{i}(\Omega_{i},a)).
\end{equation}
For every step, the agent selects its action based on the max value of these actions with an epsilon-greedy exploration strategy.

As Q-Table can not handle high dimensional states, Deep Q-Learning (DQN) uses neural network to approximate Q-function. DQN tries to optimize $Q^*_{i}$ with back propagation by minimizing the follwing loss function:
\begin{equation}
    \begin{aligned}
&\mathcal{L}_{i}(\theta)=\mathbb{E}_{\Omega, a, r, \Omega^{\prime}}\left[\left(Q^{*}_{i}(\Omega, a \mid \theta)-y\right)^{2}\right], \\
&\text { where } y=r+\gamma \max _{a^{\prime}} \bar{Q}^{*}_{i}\left(\Omega^{\prime}, a^{\prime}\right).
\end{aligned}
\end{equation}

In DQN \cite{mnih2015human}, there is a replay buffer to store agent information. Data is randomly extracted from the replay buffer to train the model. Target Q-network $\bar{Q}^{*}_{i}$ has the same architecture with ${Q}^{*}_{i}$. $\bar{Q}^{*}_{i}$ copies parameters from ${Q}^{*}_{i}$ frequently. In this way, DQN disrupts the temporal correlation and makes the training of the model more stable.

\begin{figure*}[h]
  \centering
  \label{fig:overview}
  \includegraphics[width=1.0\linewidth]{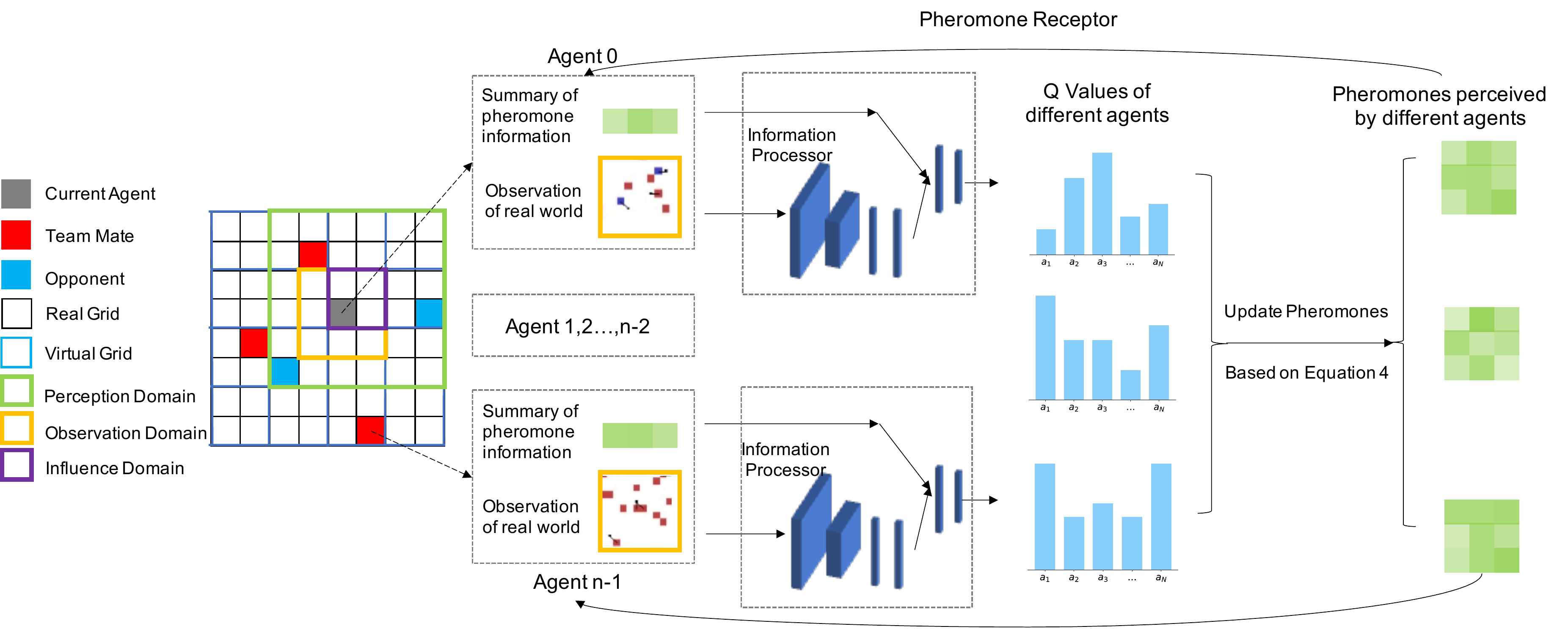}
  \caption{Overview of PooL. The left part represents agents' environment from the view of our proposed framework. The right part shows how agents process information from two sources (real world observations and pheromones) and release pheromones based on their output.}
  \label{fig:overview}
  \Description{Overview of our framework}
\end{figure*}

\section{Methodology}

In this section, we discuss how our pheromone-based framework is realized. Firstly, a general view of PooL is described. Then, implementation details will be introduced step by step. We choose Q-Learning as a base model for PooL in the following sections but our framework is able to combine with most existing reinforcement learning algorithms.

\subsection{Communication Framework Overview}

Figure \ref{fig:overview} gives a whole picture of PooL. In the left of Figure \ref{fig:overview}, a simple grid world is built as a demonstration. From the view of the grey agent, the black-bordered squares represent the real grids of our environment. They are the smallest units to express information in our environment. In order to realize indirect communication based on pheromones, we need to set up a virtual medium, which is represented by the blue-bordered squares. The area near the current agent is divided into three regions from small to large: Influence Domain, Observation Domain, and Perception Domain. Agents handle these three regions with different behaviors. The meanings of the three regions are as follows.

\begin{itemize}
  \item \textbf{Influence Domain}: the influence range of pheromones released by agents. 
  \item \textbf{Observation Domain}: the partial observation of the agent in the environment.
  \item \textbf{Perception Domain}: the range of pheromones that agents can perceive.
\end{itemize}

In nature, generally, animals perceive a much larger range of pheromones than they observe their surrounding environment by eyes. Therefore, in our settings, the Perception Domain is larger than the Observation Domain.

The right part of Figure \ref{fig:overview} shows how PooL processes received information and updates pheromones. PooL consists of three components. The first is the information processor (composed of convolution layers and fully connected layers). Its input feature consists of two parts: real-world observations and a summary of pheromone information. Its output is Q-values, which can be transformed into pheromones. The second is the pheromone update mechanism realized through a virtual medium. Thirdly, a pheromone receptor is used to sense the pheromone information from the surrounding by convolution layers and transform the information into the representation features of the current environment. With the increase of the number of agents, the pheromone information extracted by agents can more accurately reflect the situation of the local environment.

In the following sections, we introduce the realization of our framework in three steps. First, a virtual medium is defined to allow the propagation of pheromones. Then, when agents release their pheromones, an update mechanism updates the values of pheromones stored in the virtual medium. Finally, pheromone information is processed by the pheromone receptor and serves as extra input for our reinforcement learning algorithms such as DQN.

\subsection{Virtual Medium Settings}
In order to achieve pheromone-based indirect communication in the environment, a virtual medium to simulate the mechanism of pheromones in nature needs to be built first. A virtual map can be constructed by dividing the real environment map into $H \times W$ virtual grids. $pos(i)$ denotes a mapping function that can return agent $i$'s current position like $(h,w)$. $Info$ represents the pheromone value stored in the virtual medium whose shape is $(H, W, N_A)$ ($N_A$ is denoted as the number of actions). $N(h,w)$ represents the current number of agents in $(h,w)$. The pheromone released by agent $i$ is denoted as $Phe(i)$. $Phe(i)$ is a vector with $N_A$ dimensions.

\subsection{Pheromone Update Rule}

Following Q-Learning's idea, pheromones are defined based on the Q-function. As the range of the Q-function's estimated values changes while training, Q-values are standardized as pheromones. For agent $i$'s Q-values $Q_i$, they are transformed by a standardization function $St(Q_{i})$. After standardization, values of pheromones obey the standard normal distribution with the mean of $0.0$ and the standard deviation of $1.0$.

\begin{equation}
\label{eq:p_update}
\begin{aligned}
    &Phe(i)=St(Q_{i}), \\
    &\text{where } Q_{i} = (Q_{i}(\Omega_{i},a_0),\cdots,Q_{i}(\Omega_{i},a_{N_A-1})).
\end{aligned}
\end{equation}
At the beginning of each time step, each agent calculates their Q-values and releases their pheromones on their Influence Domain. We take the mean value of all released pheromones in each grid, and update the value of pheromones in each grid according to the evaporation coefficient $\beta$. For a virtual grid, whose position is $(h,w)$, its pheromone value is updated by the following equation:
\begin{equation}
\begin{aligned}
    \operatorname{Info}(h, w)=&(1-\beta) \operatorname{Info}(h, w) \\
    &+\beta \sum^{I_{h,w}} Q\left(s_{i}\right) / N(h, w).
\end{aligned}
\end{equation}
$I_{h,w}$ is a set of agents where for every $i \in I_{h,w}$, there is $pos(i)=(h,w)$.

\subsection{Pheromone Based Deep Q-Learning}
The main difference between our proposed pheromone-based deep Q-Learning and DQN is that our information processor receives inputs from two sources. Except for the partial observation of the current environment, our processor also receives the information processed by the pheromone receptor. For agent $i$, the summary of pheromone information processed by agent $i$ is denoted as $Rec(i, Info)$. The optimization goal for Q-learning can be updated to:
\begin{equation}
\label{eq:loss}
\begin{aligned}
&\mathcal{L}_{i}(\theta)=\mathbb{E}_{\Omega, Info, a, r, \Omega^{\prime},Info^{\prime}}\left[\left(Q^{*}_{i}(\Omega, Rec(i, Info \mid \theta_r), a \mid \theta_p)-y\right)^{2}\right], \\
&\text { where } y=r+\gamma \max _{a^{\prime}} \bar{Q}^{*}_{i}\left(\Omega^{\prime}, Rec(i, Info^{{\prime}}),a^{\prime}\right).
\end{aligned}
\end{equation}

The complete algorithm is described in Algorithm \ref{algo:1}. The algorithm can be divided into two parts: calculate and update pheromones released by different agents and choose actions based on information observed by agents. 

\begin{algorithm}
\renewcommand{\algorithmicrequire}{\textbf{Input:}}
\renewcommand{\algorithmicensure}{\textbf{Output:}}
\caption{PooL Motivated Deep Q-Learning}
\label{algo:1}
\begin{algorithmic}[]
\REQUIRE $\mathbb{D}$: replay buffer; $\theta^0_p,\theta^0_r$: initial parameters of information processor $Q$ and pheromone receptor $Rec$; $T$: maximum number of steps allowed to run the game; $N_{e}$: maximum number of running episodes; $Info$: an all zero matrix of shape $H \times W \times N_A$. 
\ENSURE Optimal parameters  $\theta^*_p,\theta^*_r$ for trained model.
\FOR{round $r$ = $1$ to $N_{e}$}
    \FOR{step $t$ = $1$ to $T$}
        \FOR{all $(h,w)$ satisfies $1 \leq h \leq H$ and $1 \leq w \leq W$}
        \STATE update $Info(h,w)$ with equation \ref{eq:p_update}
        \ENDFOR
        \STATE Select Action $a_i$ for every agent by $a_i=\mathop{\arg\max}_{a} Q_{i}(\Omega, Rec(i, Info|\theta_r), a \mid \theta_p)$
        \STATE  Update the environment state according to the actions taken by agents.
        \STATE For every agent, store $(\Omega_i,a_i,r_i,Info,\Omega^{\prime},Info^{\prime})$ to $\mathbb{D}$
    \ENDFOR
\STATE Sample a mini-batch data from $\mathbb{D}$
    \STATE Calculate loss of the data by Equation \ref{eq:loss}
    \STATE Perform a gradient descent step to update $\theta^e,\theta^d$
    \STATE Copy parameters from the current network to the target network at a certain frequency
\ENDFOR
\end{algorithmic}
\end{algorithm}

\subsection{Algorithm Cost Analysis}

Compared with DQN, the extra cost of PooL lies in the pheromone update mechanism and pheromone receptor. For pheromone update, it can be done in $O(H W)$ time. For the pheromone receptor, the complexity of its network structure depends on the size of the Perception Domain, and the size of the Perception Domain is generally much smaller than the Observation Domain. As a result, the extra cost of PooL is relatively small compared with DQN.

\subsection{Implementation Details}

As shown in Figure \ref{fig:overview}, we use the structure of multi-layer convolution and multi-layer fully connected layers as our information processor to process observations and summary of pheromone information. The receptor is also a convolution structure and processes pheromone information from the virtual medium. The size of the Influence Domain is $1 \times 1$ and the size of the Perception Domain is $3 \times 3$. The selection of hyperparameters such as the virtual map size $H, W$, and the evaporation coefficient $\beta$ will be discussed in Section 5.

\section{Experiment}
In this section, we describe the settings of our experiment environments and show the performance of PooL. We first discuss the motivation example mentioned in Section 1. Then, main results compared with other methods is shown in various settings of MAgent. Finally, Battle environment is taken as an example to discuss details about our framework. 
\subsection{Motivation Experiment}

\subsubsection{Detailed Settings of Motivation Environment}
\
\newline
\indent In our motivation environment, agents get rewards of $-1.0$ at every time step and get rewards of $100.0$ if they find food. For each agent, they will exit from the environment if they fall into holes or find food. The game ends when there are no agents in the environment. In order to set up a multi-agent environment, several agents are placed in each ant nest. This problem is solved by two methods: Table Q-Learning and PooL motivated Table Q-Learning. Agents using Table Q-Learning will update their Q-functions by Equation \ref{eq:table_q}. For agents with pheromones, expect for Equation \ref{eq:table_q}, we also maintain pheromone information by Equation \ref{eq:p_update}. At every time step, agents fuse their Q-values of the current state with pheromone information according to a certain weight. In this way, we pass information from other agents to the current agent's Q-table.

\subsubsection{Motivation Result}
\begin{figure}[htbp]
\centering

\subfigure[Motivation Environment]{
\begin{minipage}[t]{0.5\linewidth}
\centering
\includegraphics[width=1.5in]{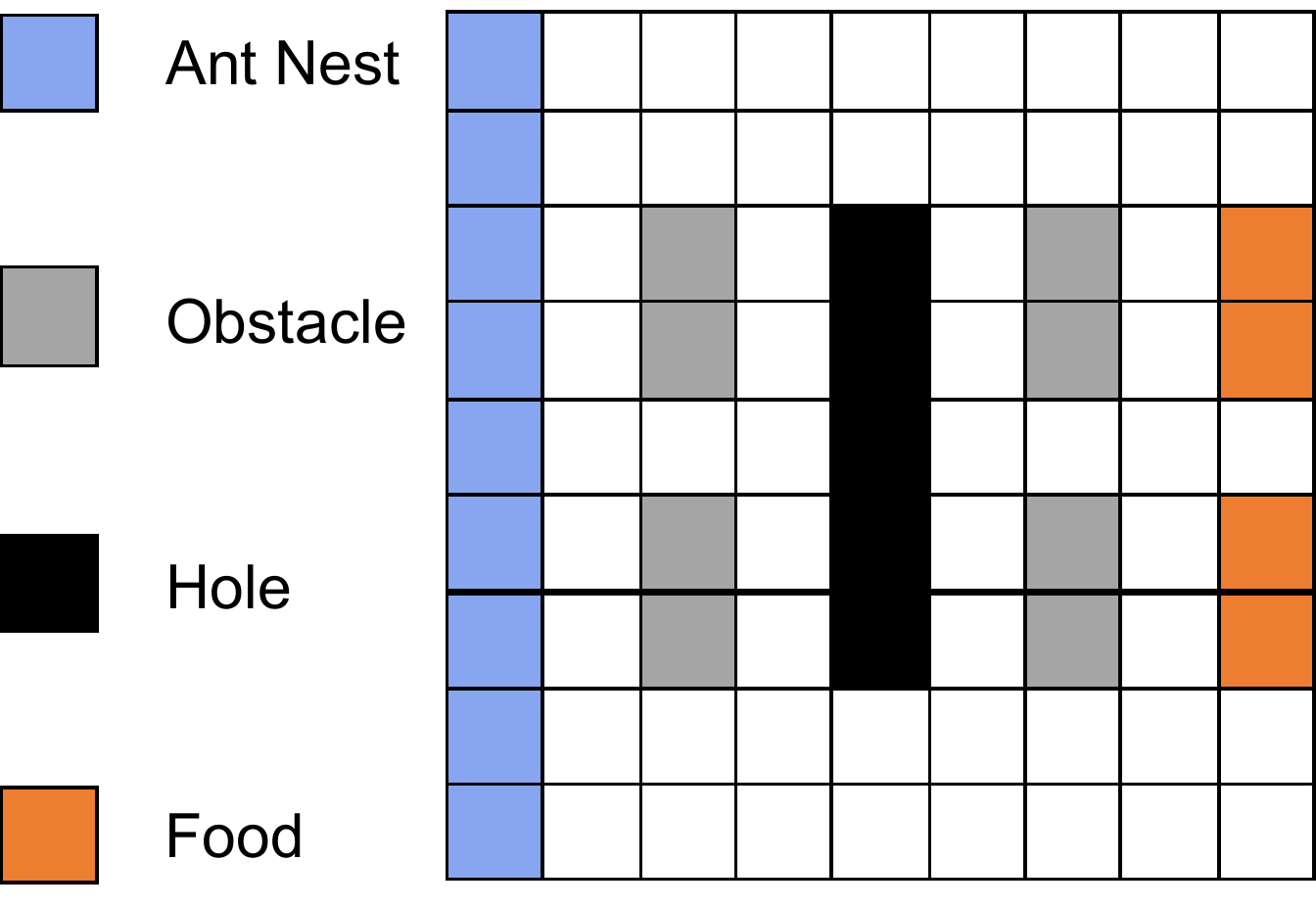}
%\caption{fig1}
\end{minipage}%
}%
\subfigure[Motivation Result]{
\begin{minipage}[t]{0.5\linewidth}
\centering
\includegraphics[width=1.8in,height=3.15cm]{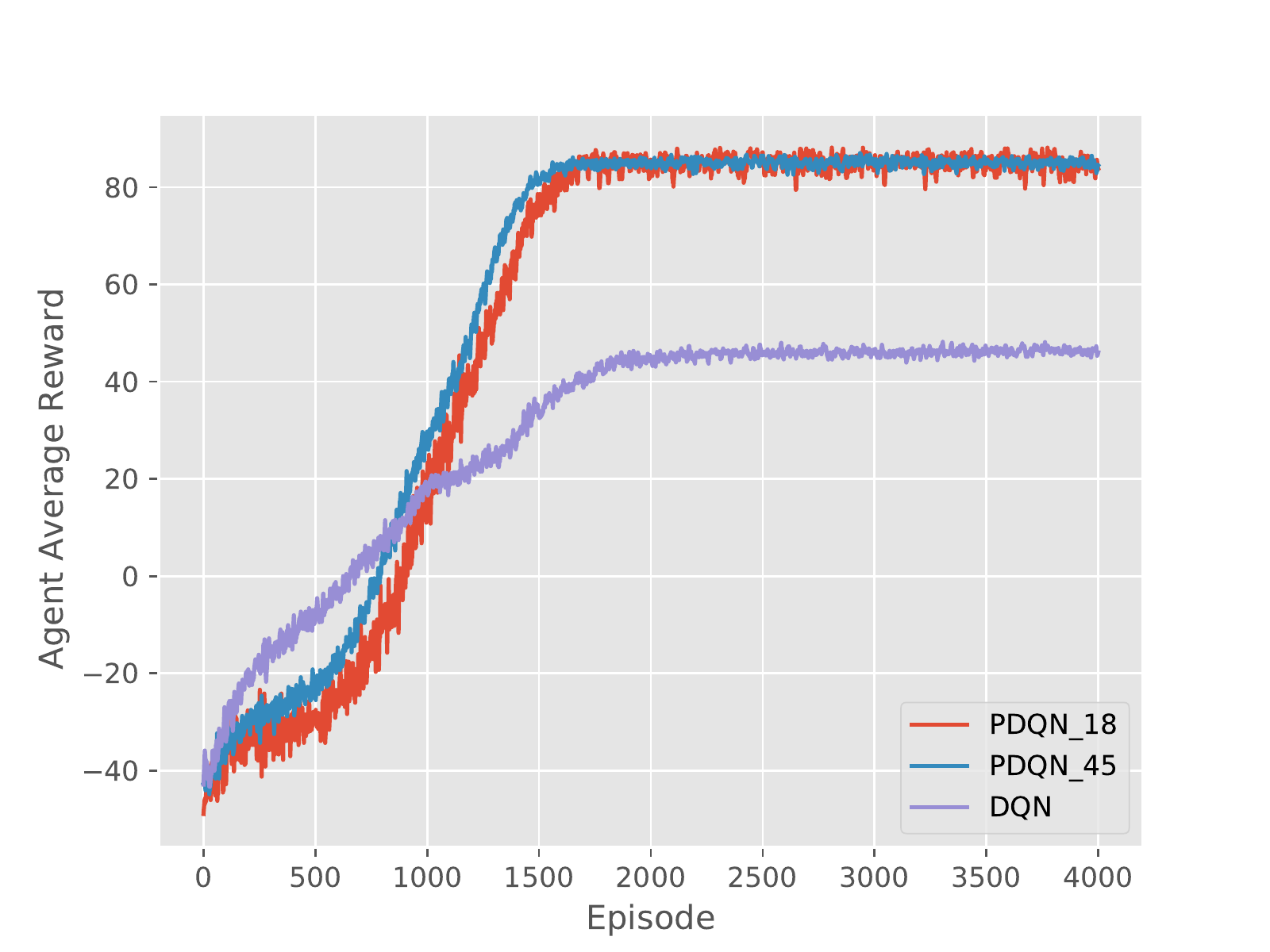}
%\caption{fig2}
\end{minipage}
}%
\centering
\caption{(a) Environment demonstration of our motivation. Agents start from ant nests on the left and try to find food. (b) Result of Motivation Experiment. Table Q-Learning (the purple line) receives lower rewards. Pool motivated Table Q-Learning (the blue line and the red line) receives higher rewards.}
\label{fig:movivation_result}
\end{figure}

\
\newline
\indent Figure \ref{fig:movivation_result} shows results of Table Q-Learning and PooL with different numbers of agents. Although Table Q-Learning achieves higher rewards at the beginning, more agents of PooL find the optimal solution at the end, which leads to higher average rewards for PooL. Our framework's rewards will be smoother if there are more agents in the environment, which indicates that our framework is more advantageous when there are more agents.

\subsection{Main Experiments}

PooL is mainly evaluated in six multi-agent environments. We first introduce the settings of these six environments, and then show performance of different methods in these six environments. 
\subsubsection{Environment Description}
\begin{figure*}[htbp]
\centering
\subfigure[Battle]{
\begin{minipage}[t]{0.1666\linewidth}
\centering
\includegraphics[width=1.1in]{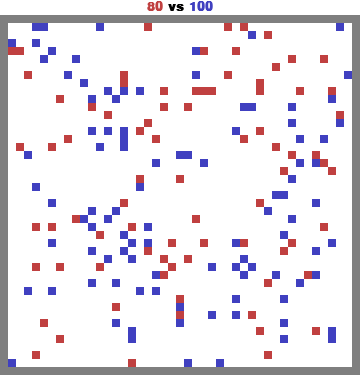}
%\caption{fig1}
\end{minipage}%
}%
\subfigure[Battlefield]{
\begin{minipage}[t]{0.1666\linewidth}
\centering
\includegraphics[width=1.1in]{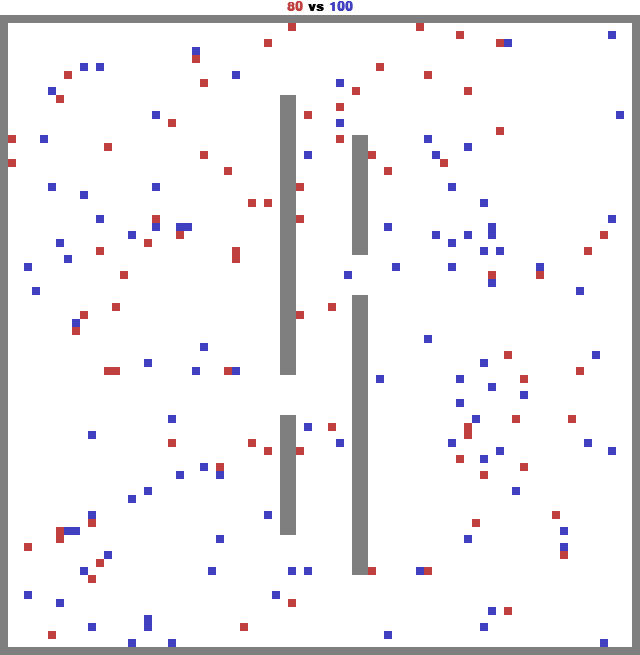}
%\caption{fig2}
\end{minipage}%
}%
\subfigure[Combined Arms]{
\begin{minipage}[t]{0.1666\linewidth}
\centering
\includegraphics[width=1.1in]{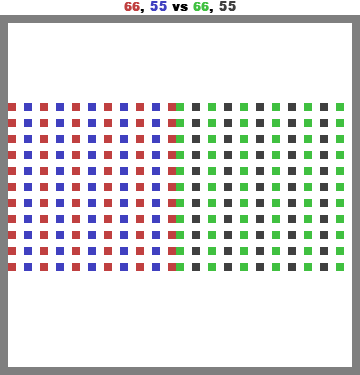}
%\caption{fig2}
\end{minipage}
}%
\subfigure[Adversarial Pursuit]{
\begin{minipage}[t]{0.1666\linewidth}
\centering
\includegraphics[width=1.1in]{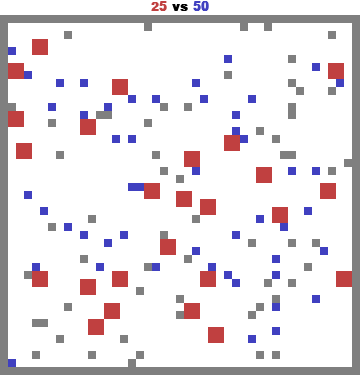}
%\caption{fig2}
\end{minipage}
}%
\subfigure[Tiger-Deer]{
\begin{minipage}[t]{0.1666\linewidth}
\centering
\includegraphics[width=1.1in]{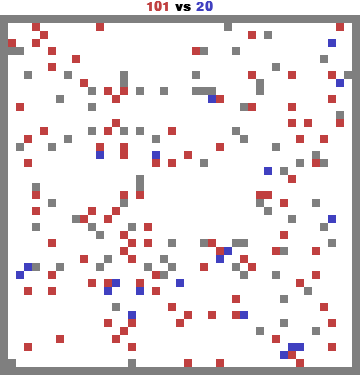}
%\caption{fig2}
\end{minipage}
}%
\subfigure[Gather]{
\begin{minipage}[t]{0.1666\linewidth}
\centering
\includegraphics[width=1.1in]{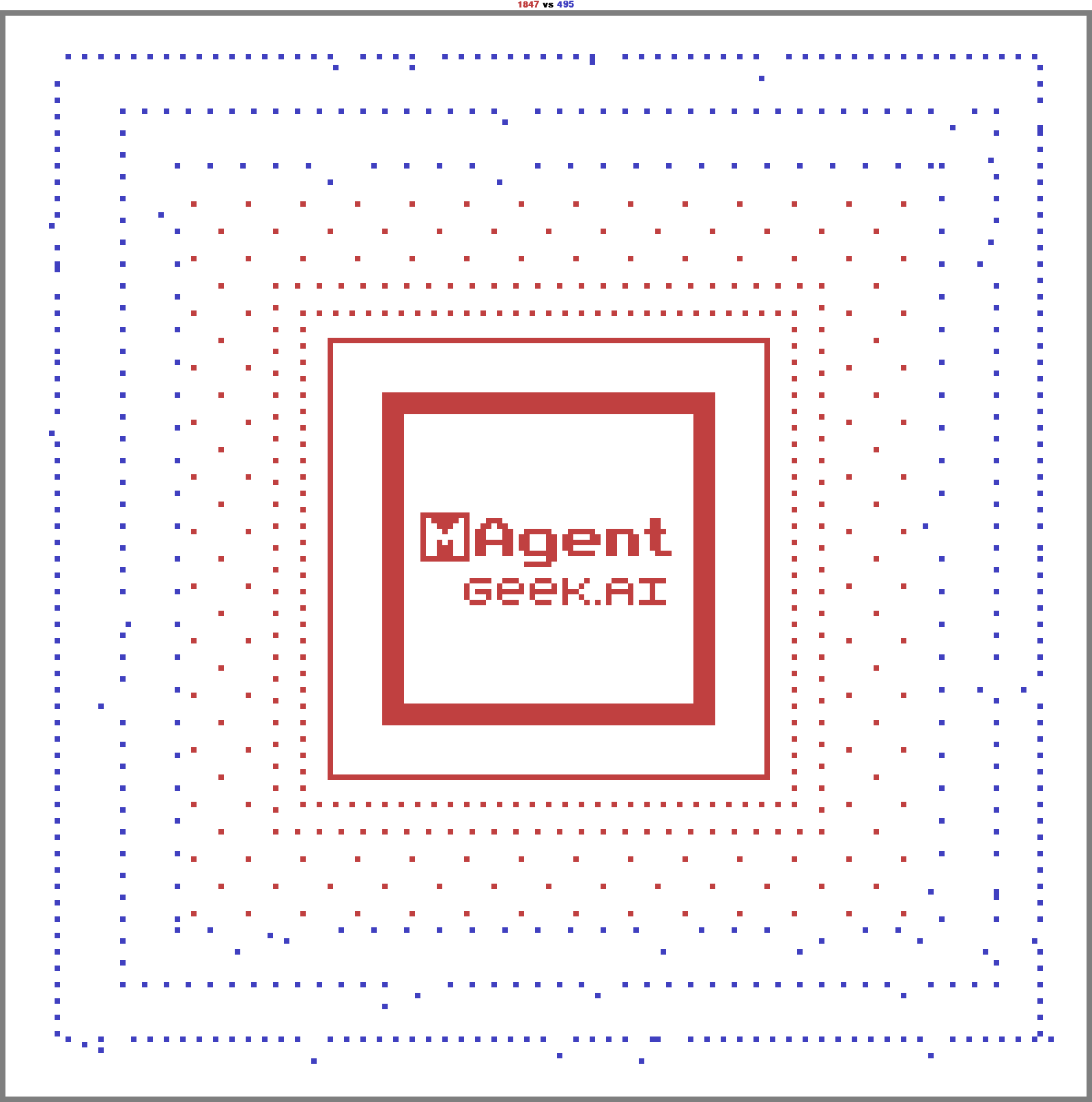}
%\caption{fig2}
\end{minipage}
}%
\centering
\caption{Demonstration of Pettingzoo MAgent enrironment.}
\end{figure*}
\
\newline
\indent 
Our environment settings are based on MAgent\cite{zheng2018magent} encapsulated by Pettingzoo\cite{terry2020pettingzoo}. Pettingzoo implements a variety of large-scale multi-agent environments based on MAgent. Agents in MAgent settings can get the location of obstacles and the location and HP (Health Points) of other agents within their view range. In our experiment, we use Battle, Battlefield, Combined Arms, Tiger-Deer, Adversarial Pursuit, and Gather as our running multi-agent environments. In the following experiments, the evaporation coefficient is $0.5$ and the virtual map size is $10 \times 10$ for Gather with real-world map size $200 \times 200$ and $8 \times 8$ for other environments with smaller real-world map sizes. 

\begin{itemize}
  \item \textbf{Battle}: In Battle, there are two teams: Red and Blue. At each time step, agents from the two teams decide their actions: move or attack. Agents will get rewards if they attack or kill their opponents.
  \item \textbf{Battlefield}: The settings of Battlefield are similar to Battle except that there are indestructible obstacles in Battlefield.
  \item \textbf{Combined Arms}: The settings of Combined Arms are also similar to Battle except that for each team there are two kinds of agents: ranged units and melee units. Ranged units have a longer attack range and are easier to be killed, while melee units have a shorter attack range and are harder to be killed.
  \item \textbf{Tiger-Deer}: Tiger-Deer is an environment that simulates hunting in nature. Tigers try to attack and kill deer to get food. Deer will die if they are attacked too many times and tigers will starve to death if they can't kill deer for a few time steps. 
  \item \textbf{Adversarial Pursuit}:Adversarial Pursuit settings are similar to Tiger-Deer without death. There are two teams in this environment: Red and Blue. Red agents are slower and larger than blue agents. Red agents will get rewards by tagging blue agents and blue agents try to avoid being tagged. 
  \item \textbf{Gather}: There are hundreds of agents and thousands of pieces of food in this environment. Agents should work together to break the food into pieces and eat them. Agents can attach each other to snatch food. It is an environment where competition and cooperation coexist. But in this paper, we focus more on cooperation. We limit the number of time steps in the game. In the case of sufficient food, agents can reduce competition and focus on cooperation.   

\end{itemize}

\begin{table*}[]
\caption{Detail settings of Experiment Environment. For Battle and Battlefield, Our agents can be either red or blue. For Combined Arms, each team contains two types of agents: Ranged and Melee.}
\label{tb:1}
\begin{tabular}{ccccccc}
\hline
\multirow{2}{*}{Environment} & \multicolumn{3}{c}{Agents to be trained}      & \multicolumn{3}{c}{Agents of opponents}       \\ \cline{2-7} 
                             & Observation Space & Amount & Role             & Observation Space & Amount & Role             \\ \hline
Battle,Battle Field          & (5,7,7)           & 80     & Red or Blue      & (5,7,7)           & 100    & Red or Blue      \\
Tiger and Deer               & (5,3,3)           & 101    & Deer             & (5,9,9)           & 20     & Tiger            \\
Pursuit                      & (5,5,5)           & 100    & Blue             & (5,6,6)           & 25     & Red              \\
Gather                       & (5,7,7)           & 495    & Predator         & -                 & 1636   & Food             \\
Combined Arms                & (7,7,7)           & 66,55  & Melee, Ranged  & (7,7,7)           & 66,55  & Melee, Ranged  \\ \hline
\end{tabular}
\end{table*}

Details of our environments are shown in Table \ref{tb:1}. Observation Space describes the shape of the features from the environment received by agents.  Amount describes the initial value of the number of agents in the environment. For Battle, Battle Field, and Combined Arms which are symmetrical, there is no difference between our agents and opponents in roles they play. In particular, in Combined Arms, each team has two kinds of heterogeneous agents. More detailed information can refer to \cite{zheng2018magent, terry2020pettingzoo}.

PooL is compared with DQN \cite{mnih2015human}, MFQ \cite{yang2018mean}, DGN \cite{jiang2020graph}. DGN is applied in Battle, Battlefield, Adversarial Pursuit, Tiger-Deer environments. In these four environments, We use the same settings as DGN —— the number of agents remains unchanged during the game. But for Combined Arms and Gather, the previous setting goes against the original intention of these two environments. Therefore, we only compare PooL with MFQ and DQN in these two environments.

\subsubsection{Main Result}
\begin{figure}[htbp]
\centering

\subfigure[Battle]{
\begin{minipage}[t]{0.5\linewidth}
\centering
\includegraphics[width=1.8in,height=2.6cm]{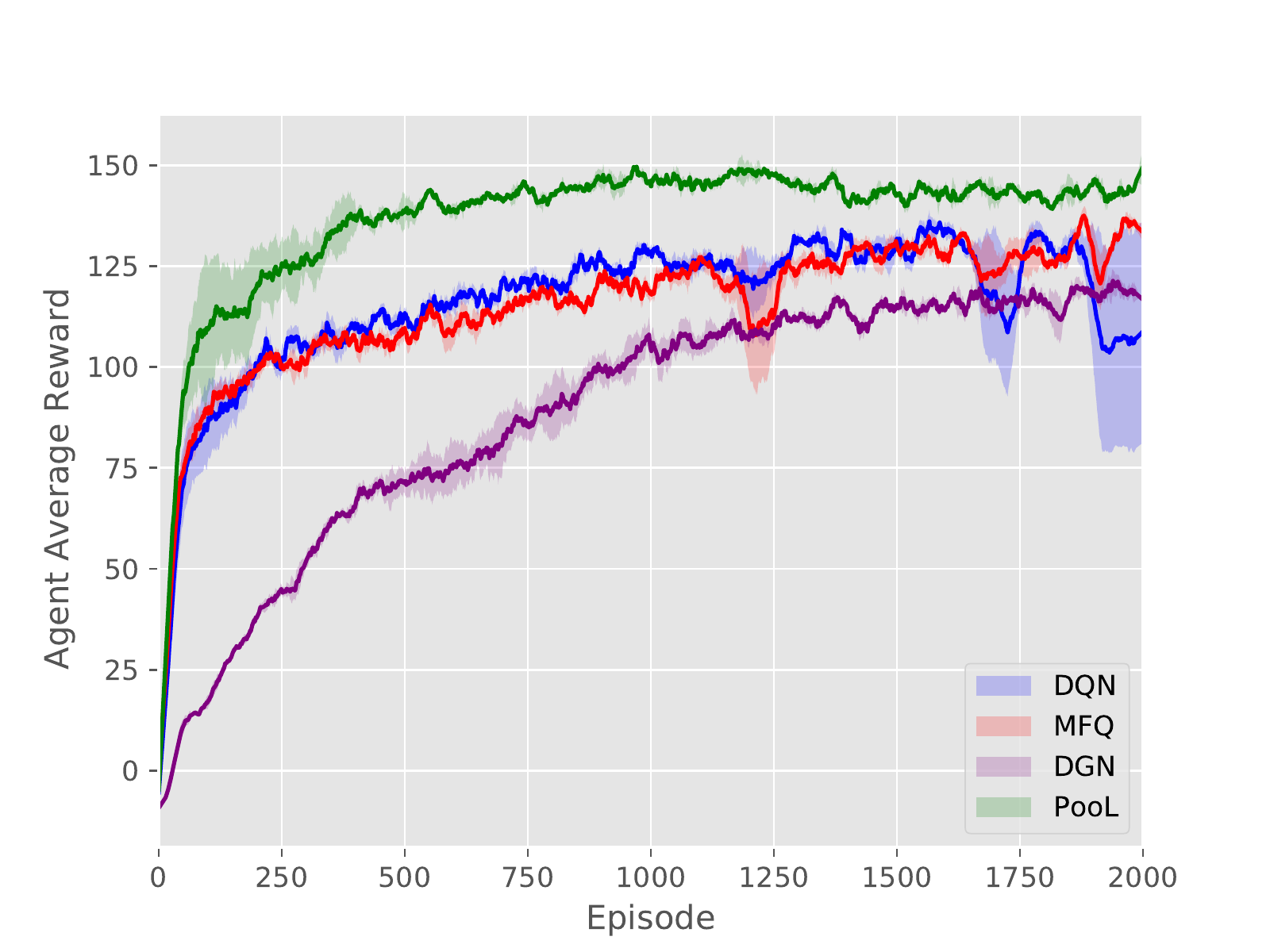}
%\caption{fig1}
\end{minipage}%
}%
\subfigure[Battlefield]{
\begin{minipage}[t]{0.5\linewidth}
\centering
\includegraphics[width=1.8in,height=2.6cm]{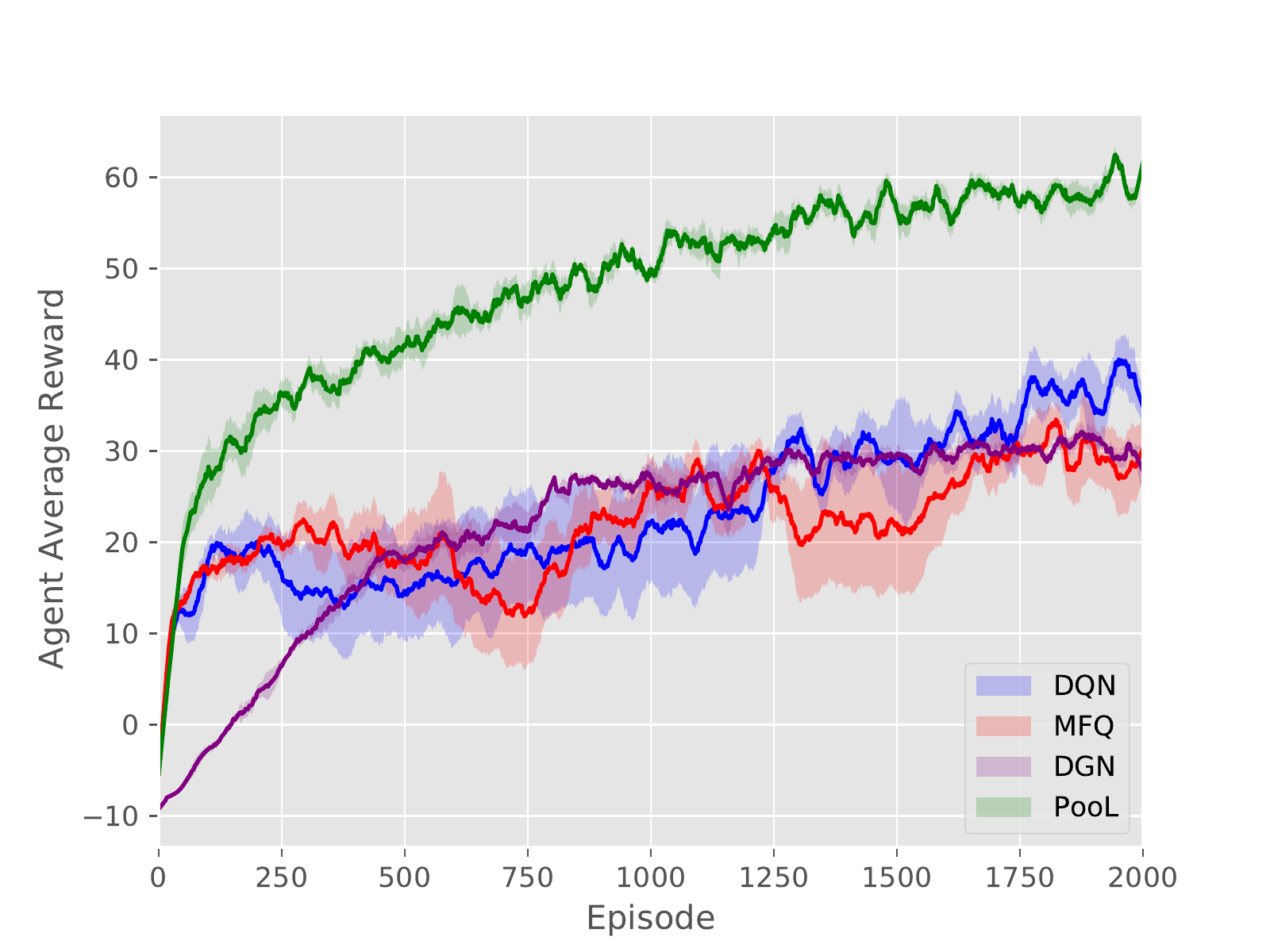}
%\caption{fig2}
\end{minipage}%
}%

\subfigure[Tiger-Deer]{
\begin{minipage}[t]{0.5\linewidth}
\centering
\includegraphics[width=1.8in,height=2.6cm]{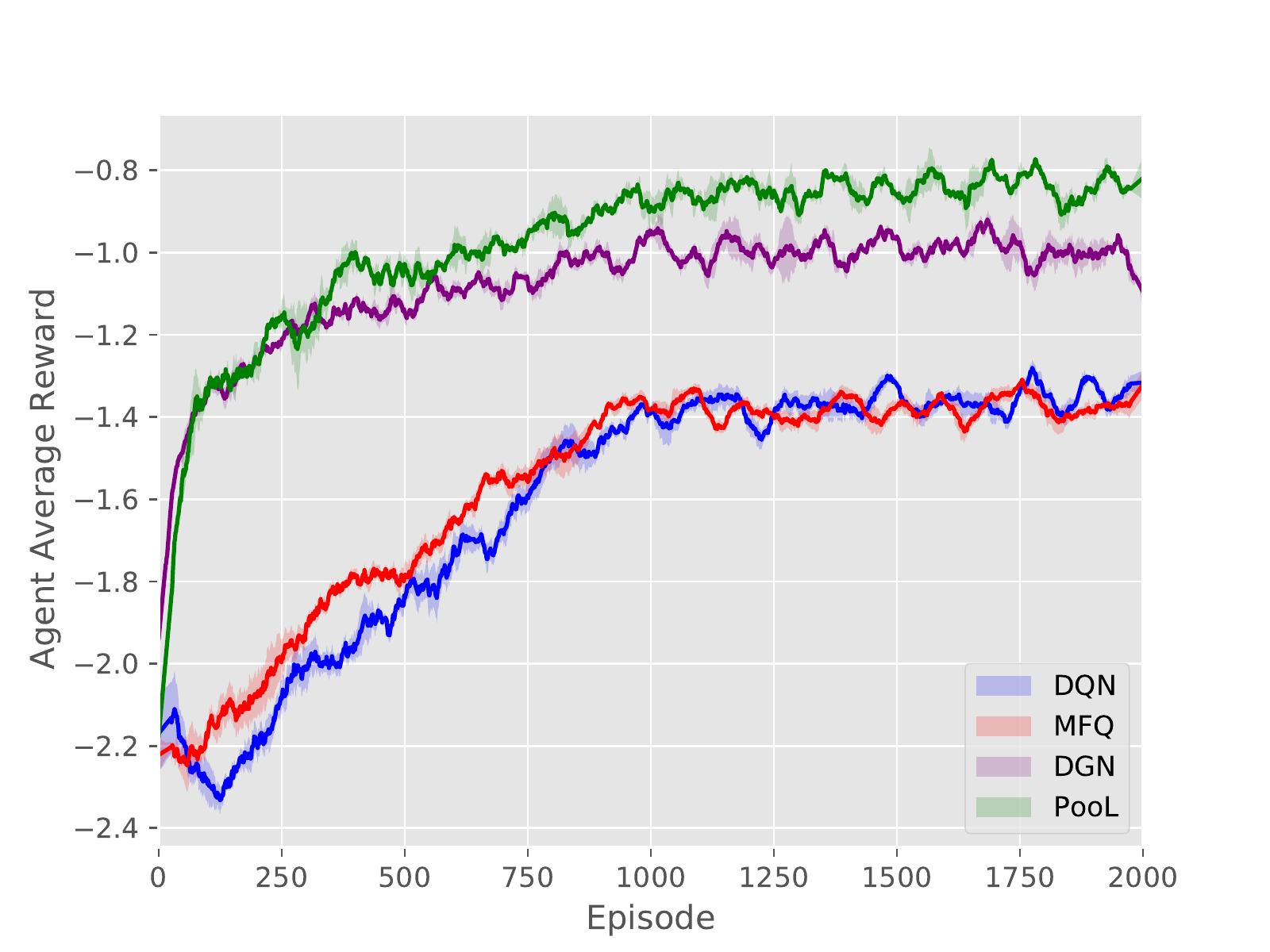}
%\caption{fig2}
\end{minipage}
}%
\subfigure[Pursuit]{
\begin{minipage}[t]{0.5\linewidth}
\centering
\includegraphics[width=1.8in,height=2.6cm]{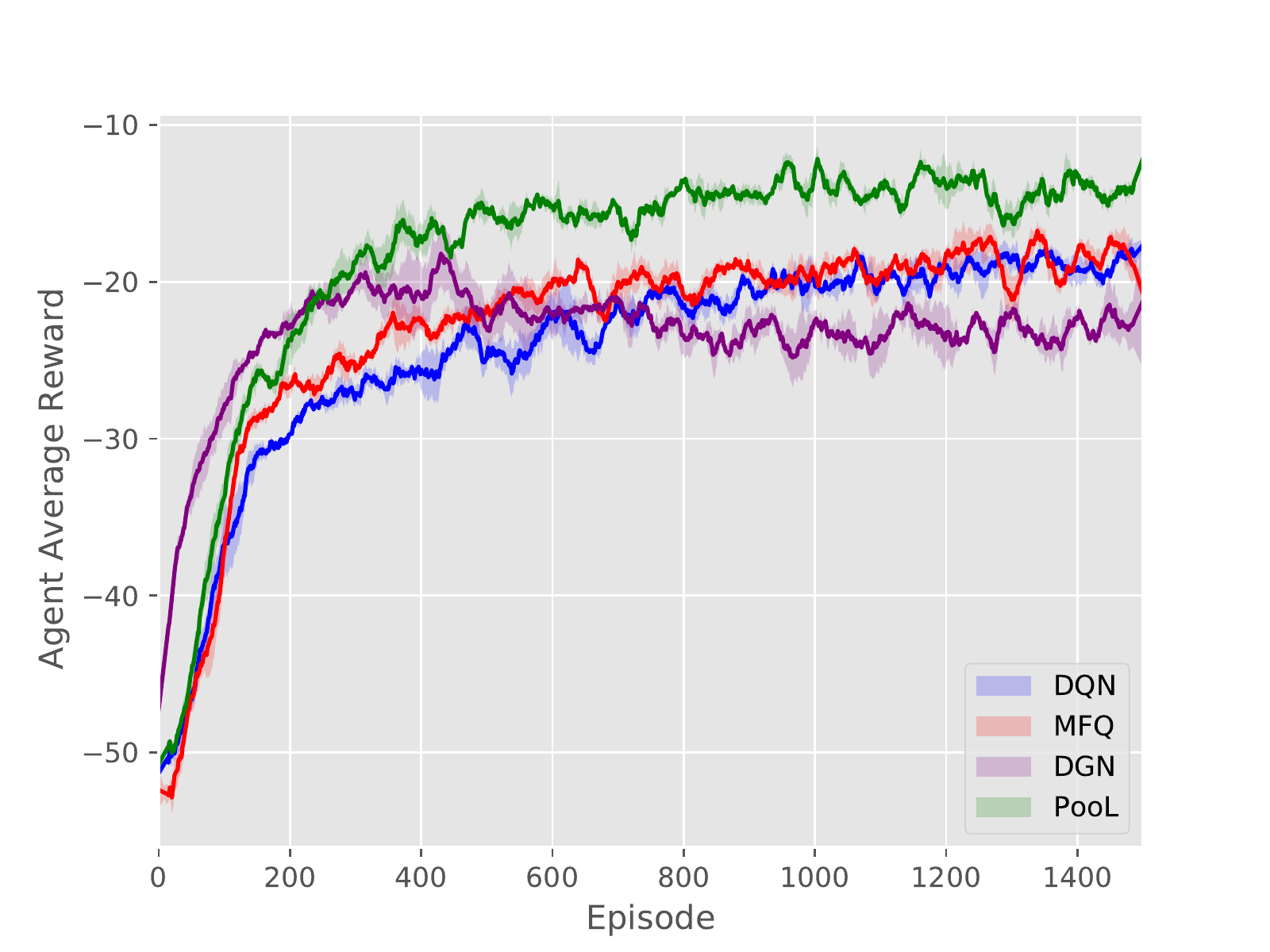}
%\caption{fig2}
\end{minipage}
}%

\subfigure[Combined Arms]{
\begin{minipage}[t]{0.5\linewidth}
\centering
\includegraphics[width=1.8in,height=2.6cm]{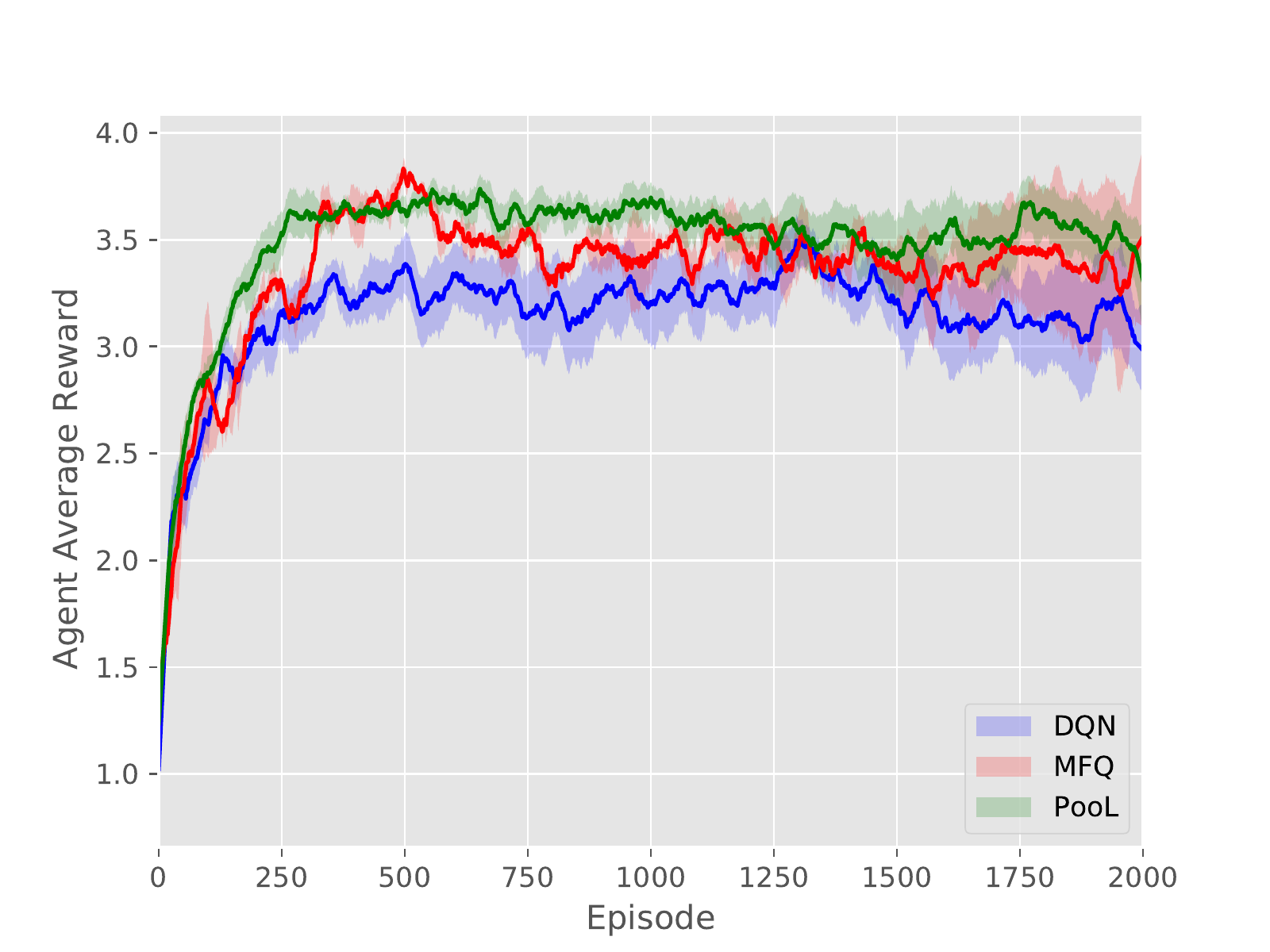}
%\caption{fig2}
\end{minipage}
}%
\subfigure[Gather]{
\begin{minipage}[t]{0.5\linewidth}
\centering
\includegraphics[width=1.8in,height=2.6cm]{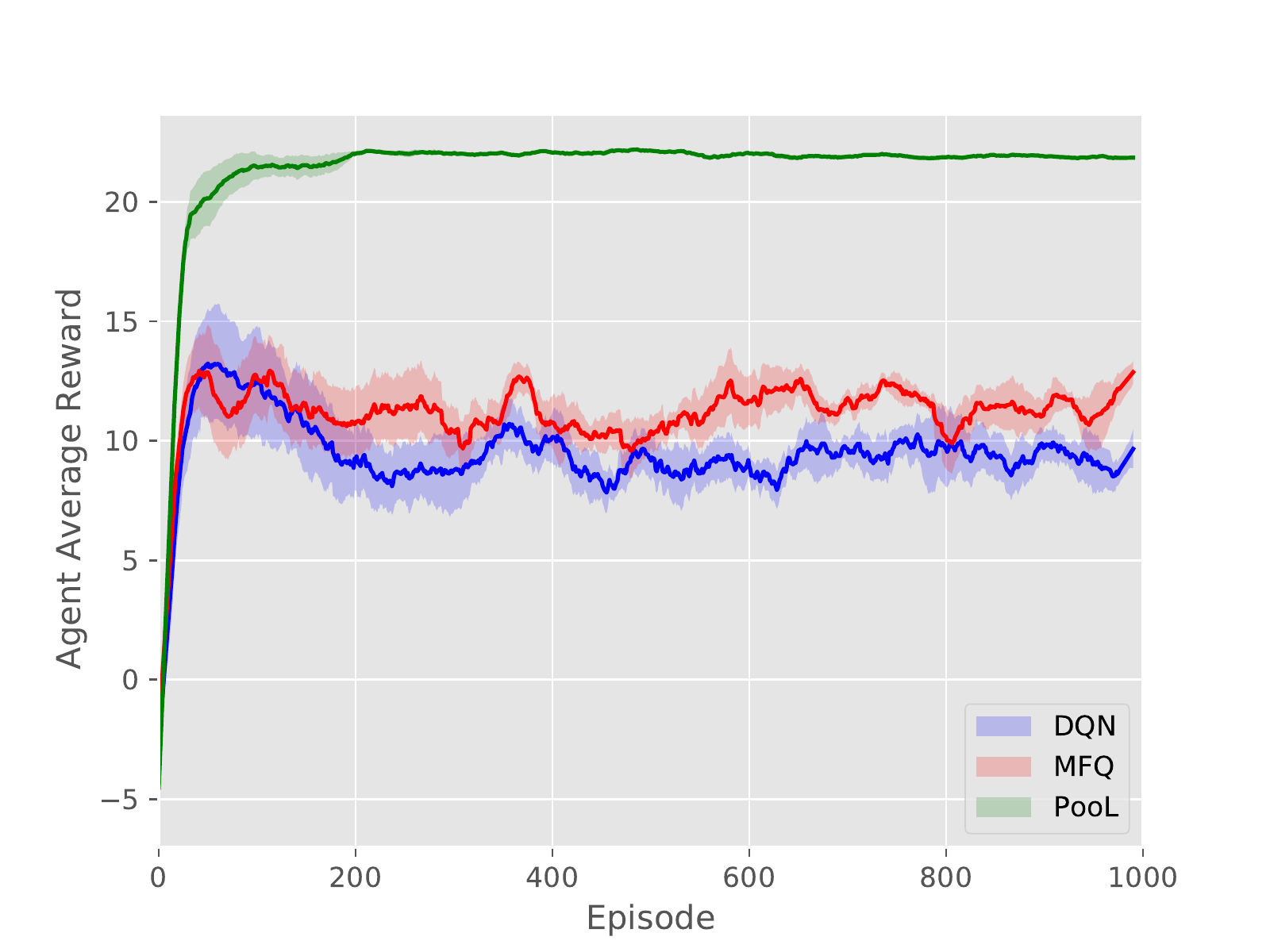}
%\caption{fig2}
\end{minipage}
}%
\centering
\caption{Main results of experiments. PooL (the green line) achieves higher rewards except in Combined Arms. The experimental results also show that PooL is more stable in repeated experiments.}
\label{fig:main}
\end{figure}

\
\newline
\indent DQN models are trained by self-play as our opponents. Then our models are trained against those opponents. In the following experiments, for PooL, the Influence Domain is $1 \times 1$ virtual grid, The Perception Domain is $3\times3$ virtual grids. The Observation Domain for different environments is shown in Table \ref{tb:1}.  Our experimental results are shown in Figure \ref{fig:main}.

\begin{itemize}
  \item \textbf{Battle}: In Battle, to improve the difficulty of our game, the team to be trained controls 80 agents and the team of opponents controls 100 agents respectively. The agents based on PooL learn the local coordination strategy faster, so as to obtain higher rewards with fewer games. 
  \item \textbf{Battlefield}: In Battlefield, the obstacles in the environment separate different agents, so for agents with only partial observations, communication is more important. As a result, PooL performs better than in Battle. With pheromones, agents based on our proposed framework obtain effective information on a larger scale and learn the coordination strategy faster.
  \item \textbf{Tiger-Deer}: We control deer to avoid being attacked by tigers. Tiger-Deer is a simulated hunting environment in nature, which is very consistent with our pheromone-based indirect communication mechanism. As a result, through pheromones, deer in the environment can quickly know which direction there is a tiger and where other deer are fleeing. 
  \item \textbf{Adversarial Pursuit}: We control the smaller blue agents to avoid being tagged by red agents. Experiments show that blue agents can communicate through pheromones to escape from red agents.
\item \textbf{Combined Arms}: As there are two kinds of agents (ranged and melee) in one team. In this experiment, our channel of $Info$ is the sum of the number of actions of the two.  Experiments show that PooL and MFQ perform better than DQN.
  \item \textbf{Gather}: In such an agent-intensive environment, experiments show that our framework can greatly promote cooperation among agents. The average reward of agents trained by PooL is about twice that of other methods.

\end{itemize}

It can be seen from Figure \ref{fig:main} that our method has a higher reward and faster convergence speed under various environmental settings. In Table \ref{tabel:2}, we choose the best model in the training step for each method and do a final evaluation. In most environments, rewards achieved by PooL are far more higher than other methods. 
% 是否需要后续进一步阐释说明

In summary, MFQ pays more attention to the coordination of global information. As a result, when the starting position of the agent is random, the performance of MFQ is similar to DQN. DGN is powerful to utilize information from its neighborhood. But as it regards agents as a graph, the convergence speed of agent training is much slower than other methods. In order to make DGN perform better in the scene of large-scale agents, we improve the update frequency of DGN parameters, which makes the time cost of DGN much higher than other methods. PooL effectively simplifies complex information in the environment, so that the agent trained by PooL can not only converge quickly but also use additional information to find better policies. PooL also has its limitation. The results in Combined Arms show that the current framework can not obtain significant advantages compared with other algorithms when there are heterogeneous agents in the same team. 
% 还差DGN，Battle的PooL
\begin{table*}[]
\caption{Evaluation Results With Trained Models. PooL ourperforms all other methods except in Combined Arms.}
\label{tabel:2}
\begin{tabular}{ccccc}
\hline
\multirow{2}{*}{Environment} & \multicolumn{4}{c}{Average Cumulated Reward} \\ \cline{2-5} 
                             & DQN        & MFQ        & DGN       & PooL               \\ \hline
Battle                       & $128.7\pm13.0$      & $124.7\pm5.4$      & $111.2\pm7.6$     & $\bm{148.5\pm12.6}$     \\
Battlefield                  & $43.7\pm7.1$       & $39.8\pm8.2$       & $55.9\pm7.1$      & $\bm{66.6\pm6.6}$      \\
Tiger-Deer                   & $-1.84\pm0.29$      & $-1.29\pm0.12$      & $-1.21\pm0.13$     & $\bm{-0.77\pm0.09}$     \\
Pursuit                      & $-18.8\pm4.6$      & $-17.6\pm4.1$      & $-18.5\pm3.0$     & $\bm{-13.5\pm5.56}$     \\
Combined Arms:Melee                & $2.00\pm0.19$       & $3.14\pm0.19$       & -         & $\bm{4.21\pm0.23}$      \\
Combined Arms:Ranged                & $3.65\pm0.31$       & $\bm{4.21\pm0.24}$       & -         & $3.90\pm0.30$      \\
Gather                       & $3.33\pm0.88$       & $9.57\pm1.17$       & -         & $\bm{22.0\pm0.32}$      \\ \hline
\end{tabular}
\end{table*}

In order to further demonstrate the advantages of our proposed framework, we use our model trained by PooL against models trained by other methods. Because only Battle and BattleField are completely symmetrical in settings, our comparison experiment is only carried out in these two environments.

\begin{figure}[htbp]
\centering

\subfigure[Comparison in Battle]{
\begin{minipage}[t]{0.5\linewidth}
\centering
\includegraphics[width=1.5in]{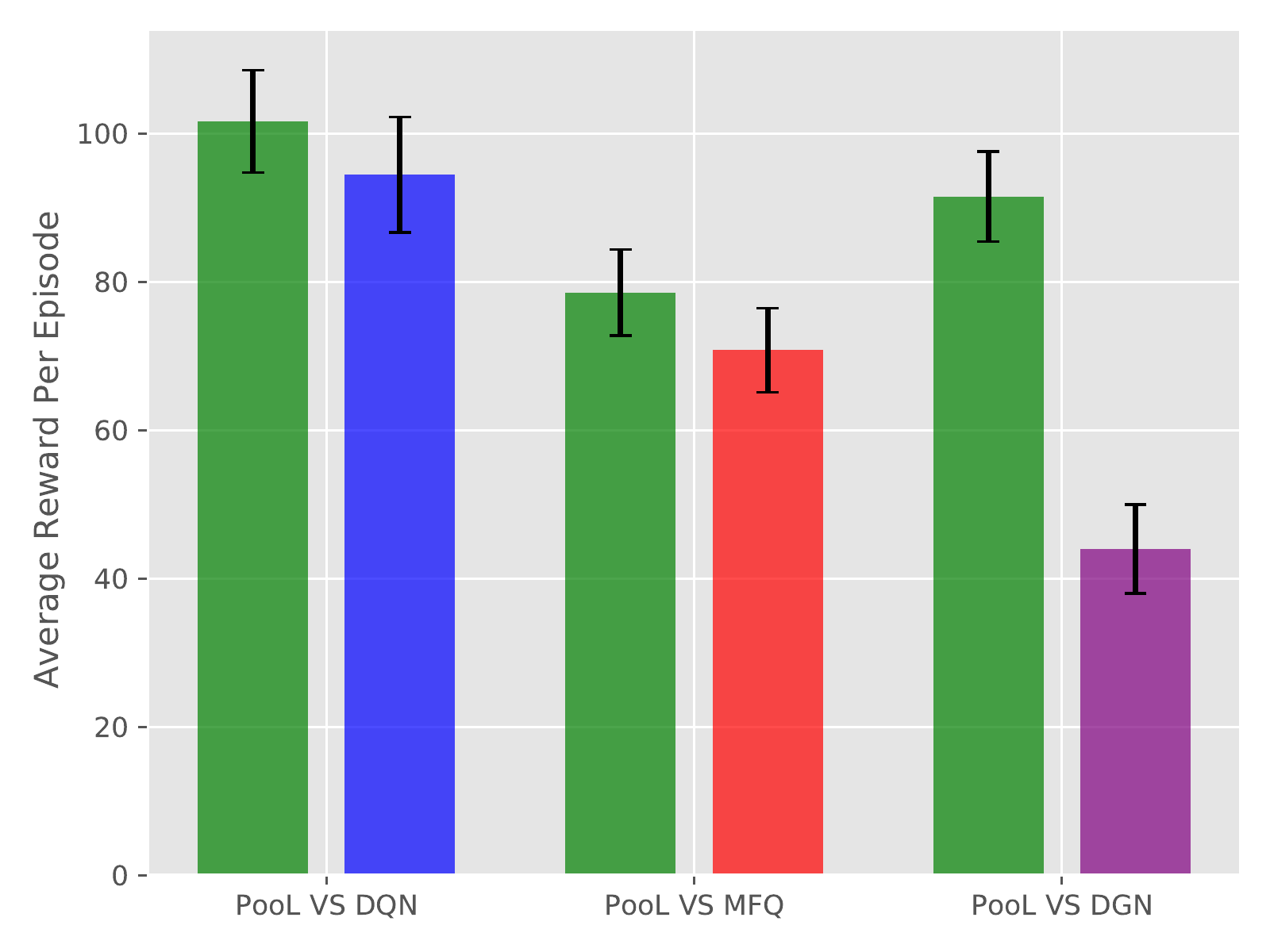}
%\caption{fig1}
\end{minipage}%
}%
\subfigure[Comparison in Battlefield]{
\begin{minipage}[t]{0.5\linewidth}
\centering
\includegraphics[width=1.5in]{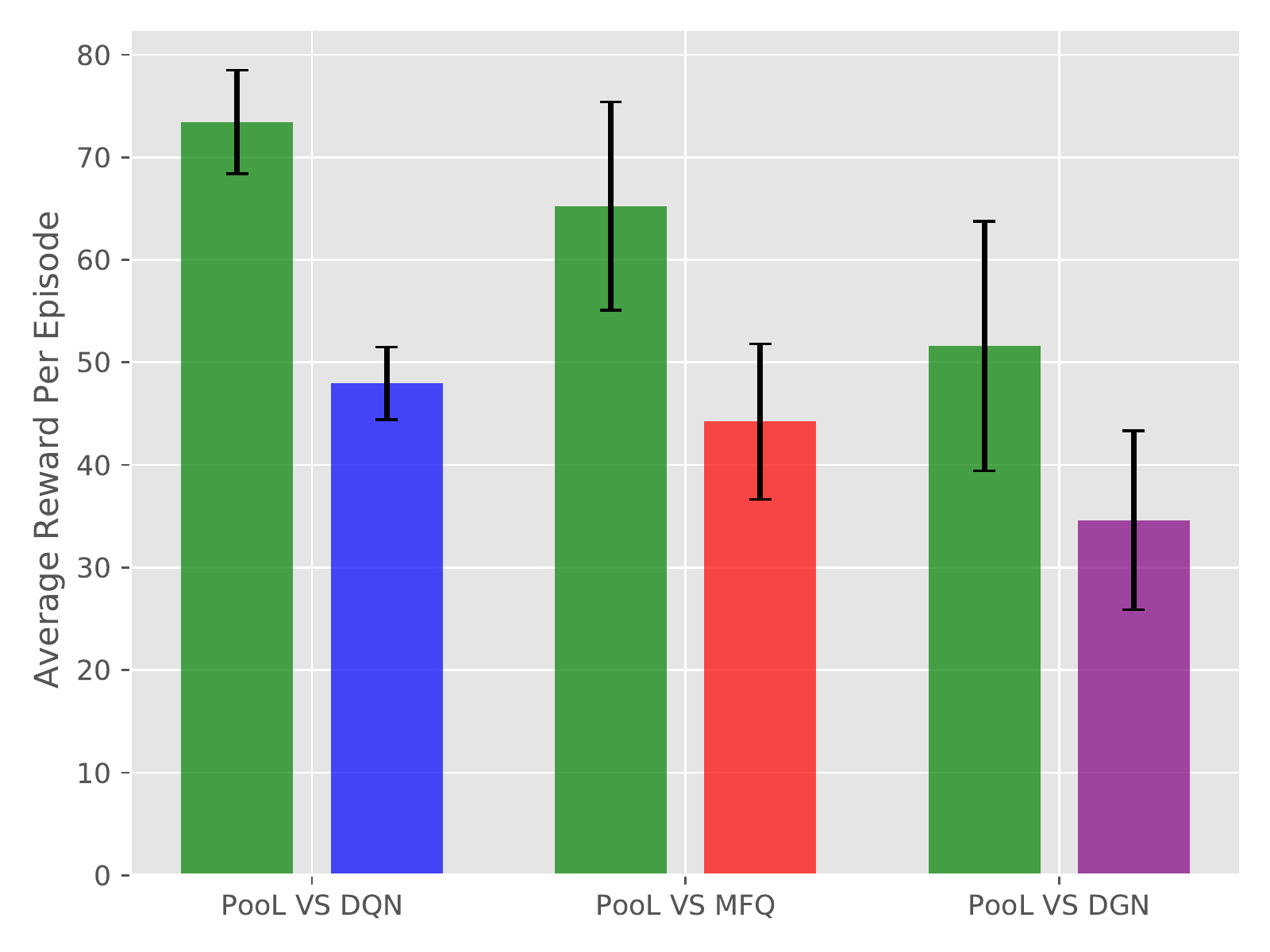}
%\caption{fig2}
\end{minipage}
}%
\centering
\caption{The performance of PooL competing against each
baseline algorithm in Battle and Battlefield.}
\label{fig:compare}
\end{figure}

Figure \ref{fig:compare} indicates that PooL not only has better convergence and cumulative reward in the training process but also has advantages against other methods. In the comparison, the overall worst performing method is DGN, which is based on graph convolution. This may be due to DGN overfitting the opponent's strategy in environments with large-scale agents. PooL not only achieves higher rewards when fighting against the baseline but also learns more general strategies than other methods.

\subsection{Detailed Experiments in Battle}

In this section, we take Battle as an example to discuss the details of our proposed method. 
\subsubsection{Effectiveness}
\
\newline
\indent In order to show the effectiveness of our method, we compare our model with DQN with global information and MFQ with neighbor information. In MAgent, the density information of agents in the environment is available. This global information is added as an extra input to DQN. MFQ method uses information from surrounding agents. In section 6.3, we follow the settings of the original source code of MFQ which averages the actions of all agents. To promote local coordination of MFQ, we divide the environment into different grids like our framework. The area of the grid is close to the area of the Perception Domain of our framework. This modified MFQ denoted as MFQ\_N receives information from its current grid.

\begin{figure}[htbp]
\centering

\subfigure[Compared with DQN with global information]{
\begin{minipage}[t]{0.5\linewidth}
\centering
\includegraphics[width=1.6in]{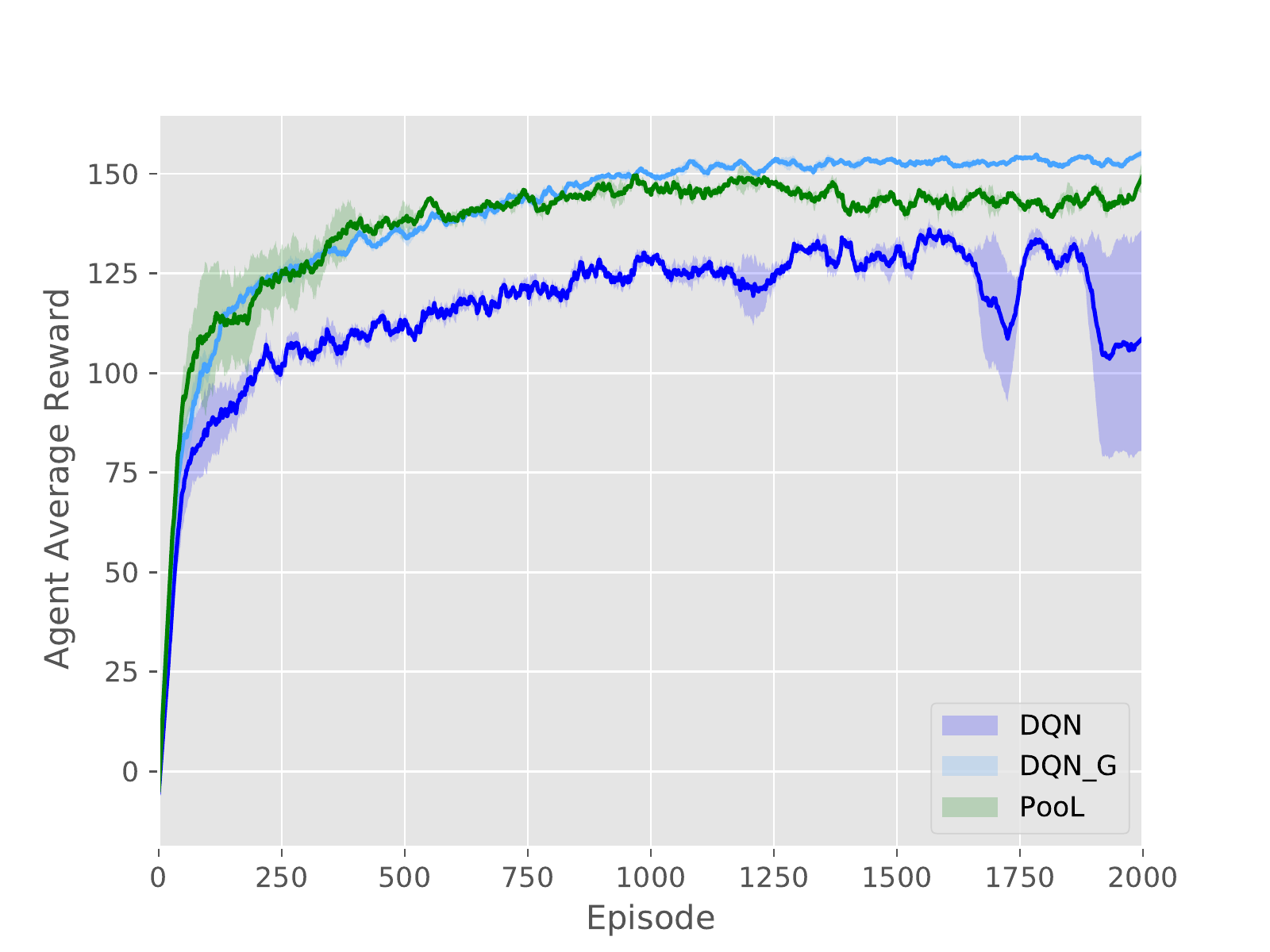}
%\caption{fig1}
\end{minipage}%
}%
\subfigure[Compared with modified MFQ]{
\begin{minipage}[t]{0.5\linewidth}
\centering
\includegraphics[width=1.6in]{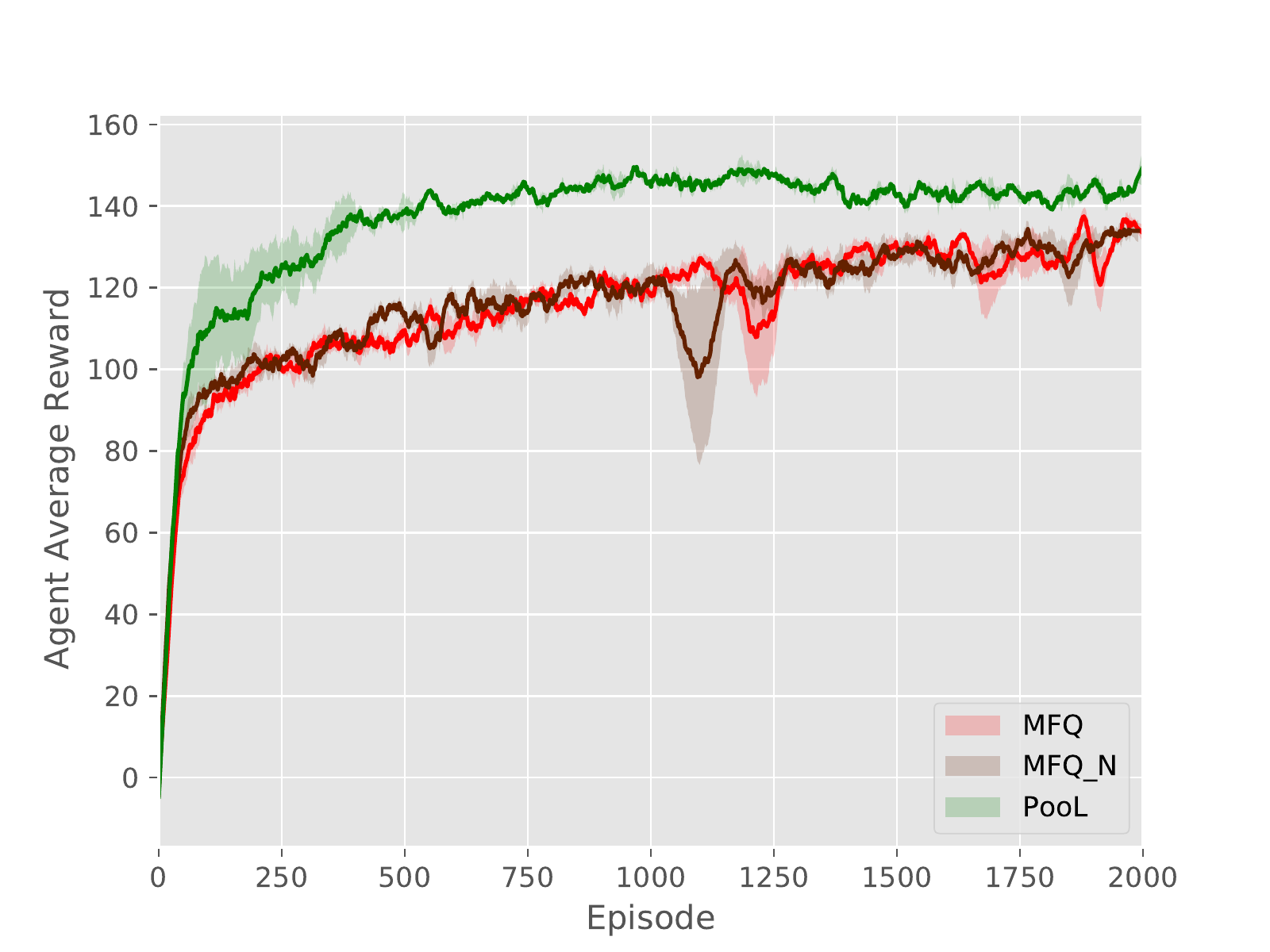}
%\caption{fig2}
\end{minipage}
}%
\centering
\caption{Performance of PooL compared with DQN with global information and MFQ with neighborhood information.}
\label{fig:effectiveness}
\Description{Performance of PooL compared with DQN with global information and MFQ with neighborhood information.}
\end{figure}

Figure \ref{fig:effectiveness} shows that our proposed framework still outperforms MFQ\_N, and achieve similar results of DQN\_G. This former result indicates that PooL not only receives signals from a wider range around but also contains more useful information than MFQ. The latter result shows that PooL is powerful to utilize local pheromone information to approximate the global situation.

\subsubsection{Scale Influence}
\begin{figure}[htbp]
\centering

\subfigure[40 VS 50]{
\begin{minipage}[t]{0.3333\linewidth}
\centering
\includegraphics[width=1.2in]{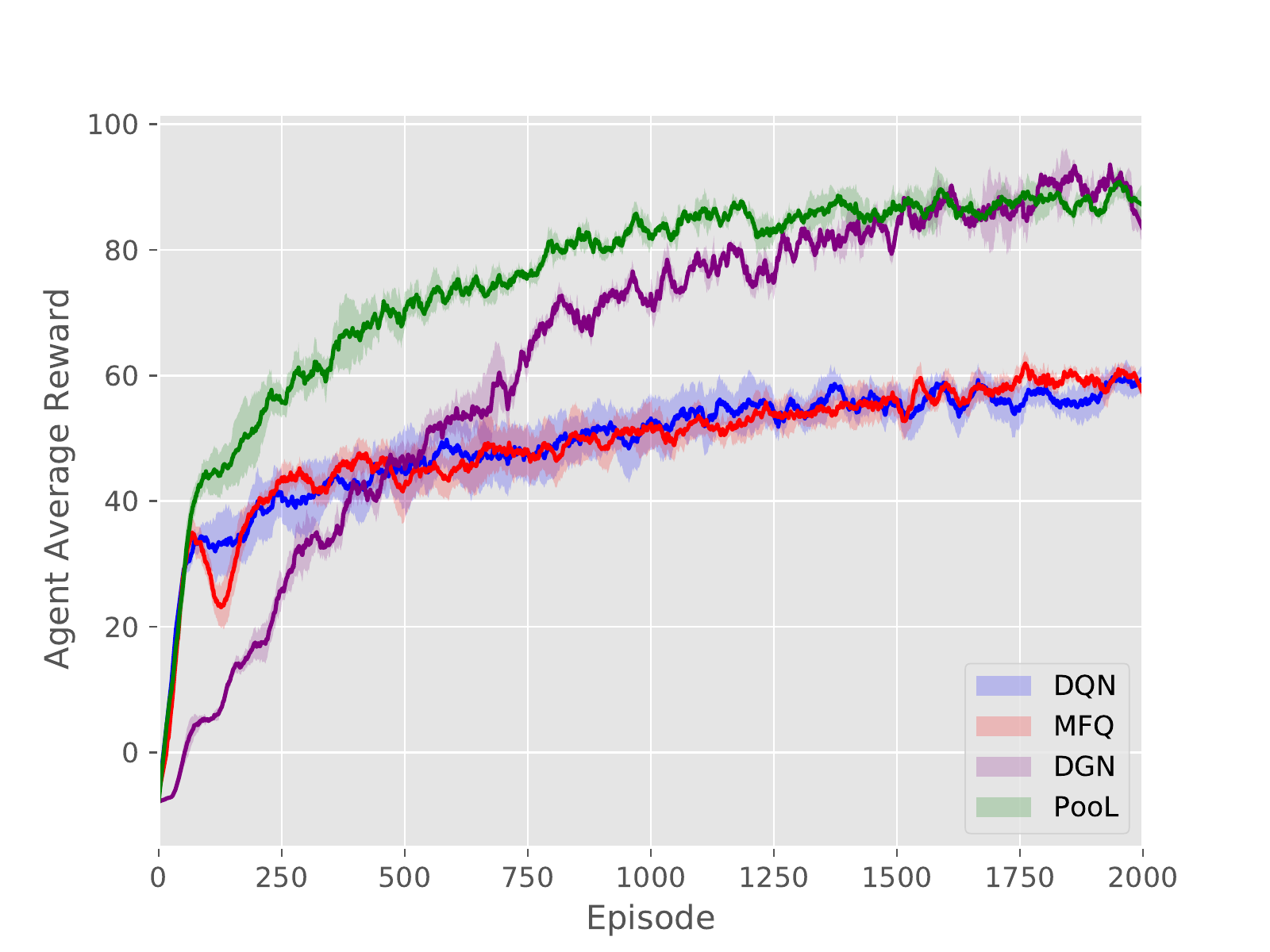}
%\caption{fig1}
\end{minipage}%
}%
\subfigure[80 VS 100]{
\begin{minipage}[t]{0.3333\linewidth}
\centering
\includegraphics[width=1.2in]{pictures/plt_battle.pdf}
%\caption{fig2}
\end{minipage}
}%
\subfigure[120 VS 150]{
\begin{minipage}[t]{0.3333\linewidth}
\centering
\includegraphics[width=1.2in]{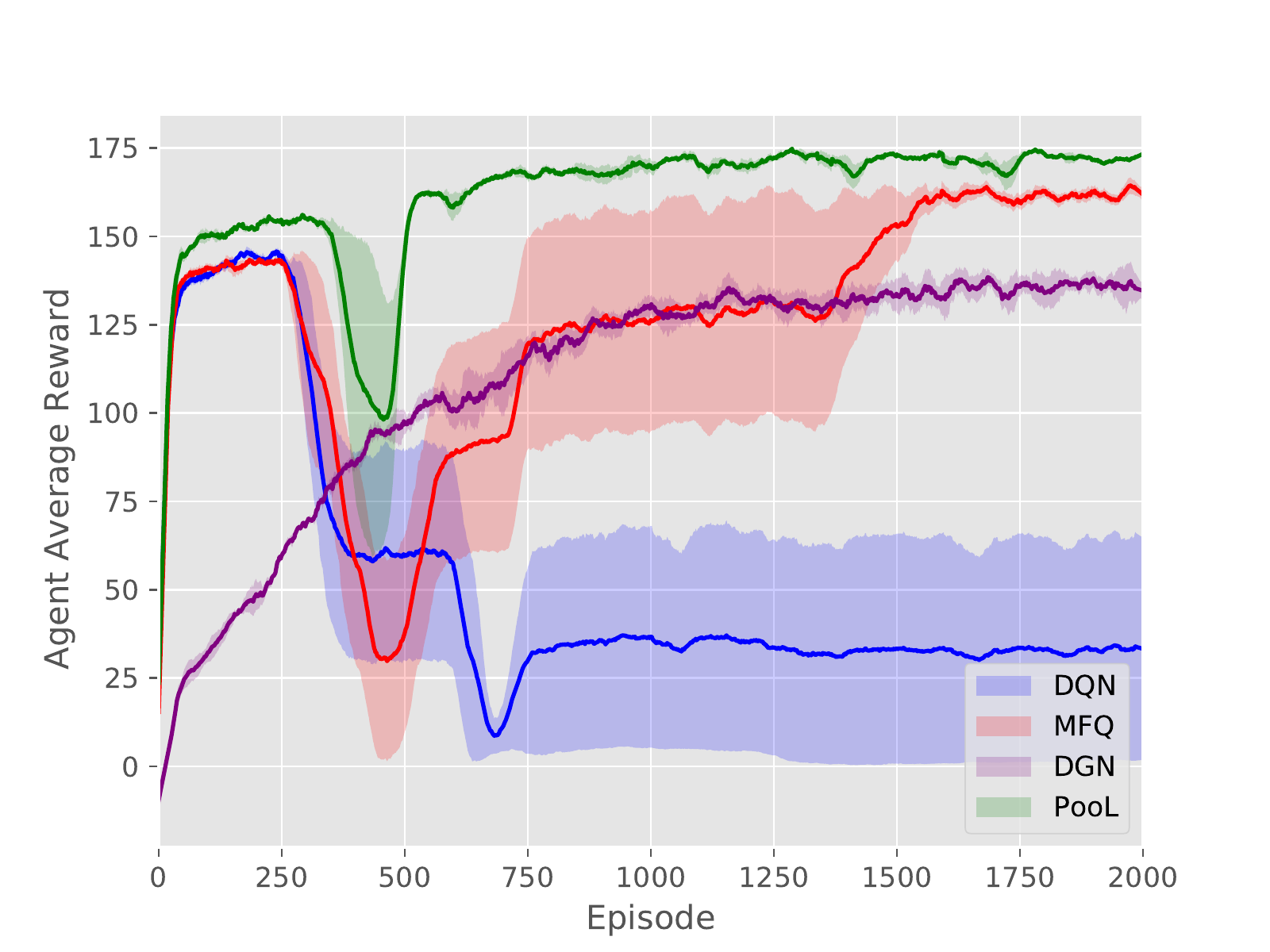}
%\caption{fig2}
\end{minipage}
}%
\centering
\caption{Scale Influence}
\label{fig:scale}
\end{figure}

\
\newline
\indent Figure \ref{fig:scale} shows how the number of agents from different teams affects our experimental result. Experimental results show that PooL is scalable and can give full play to its advantages when the number of agents in the environment is different. DQN, however, fails to learn a good policy when the number of agents becomes larger. MFQ suffers from unstable training procedures when there are more agents. 

\subsubsection{Hyper Parameters Selection}
\begin{figure}[htbp]
\centering

\subfigure[Virtual Map Size Selection]{
\begin{minipage}[t]{0.5\linewidth}
\centering
\includegraphics[width=1.6in]{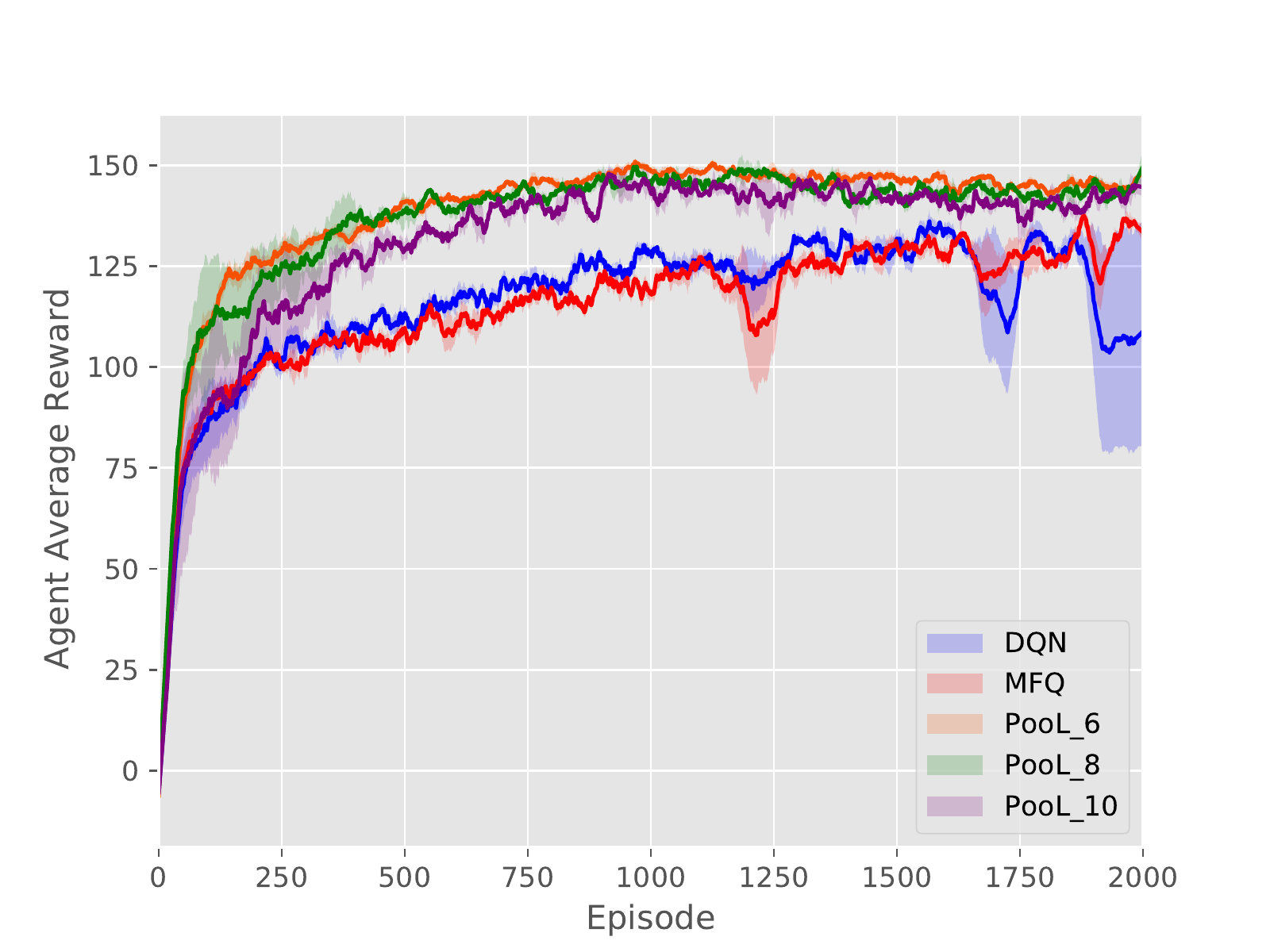}
%\caption{fig1}
\end{minipage}%
}%
\subfigure[Evaporation Coefficient Selection]{
\begin{minipage}[t]{0.5\linewidth}
\centering
\includegraphics[width=1.6in]{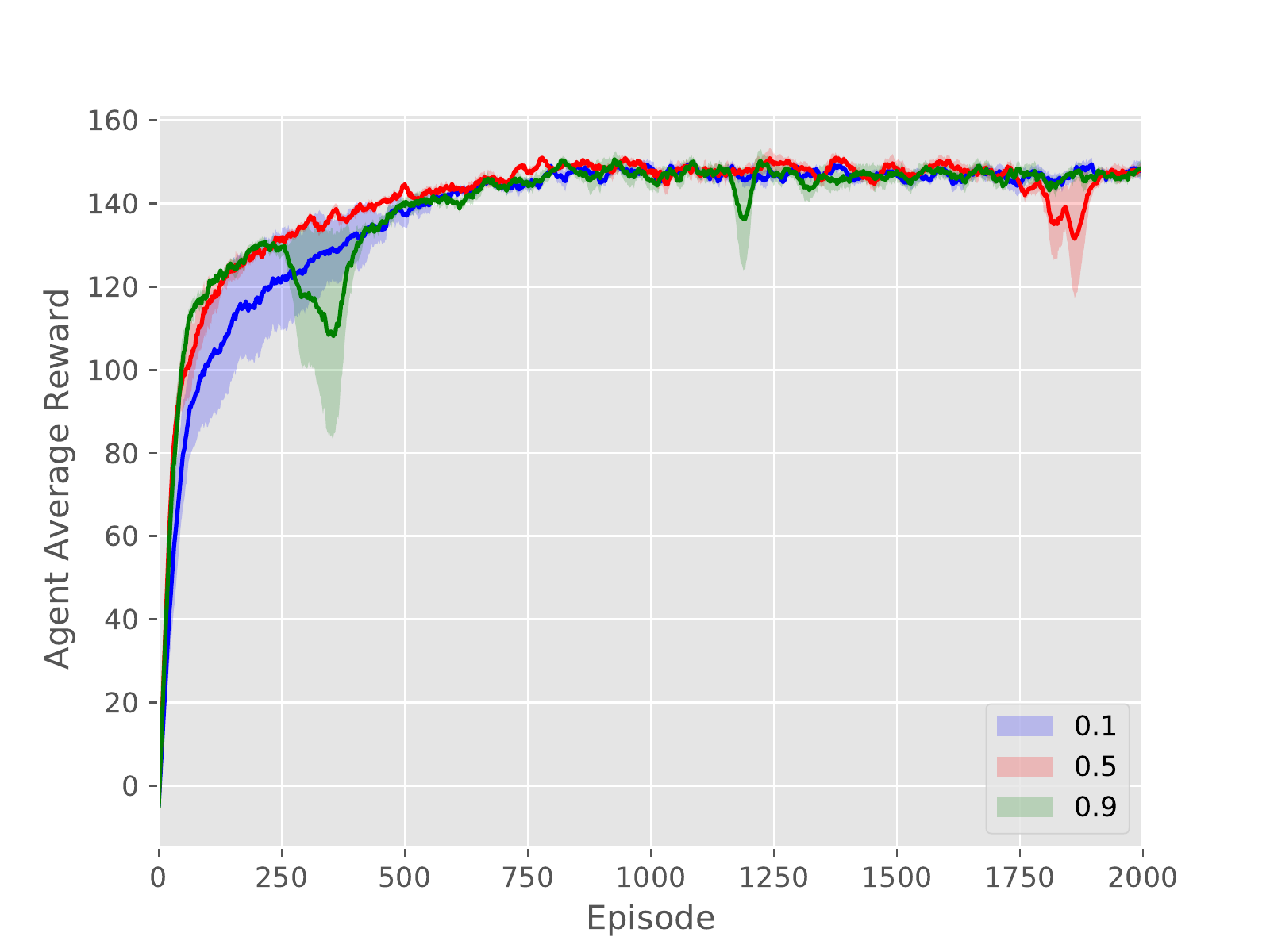}
%\caption{fig2}
\end{minipage}
}%
\centering
\caption{Hyper Parameters Selection (a) Virtual Map Size has little effect on the performance of PooL. (b) A moderate coefficient is conducive to obtaining stable training results.}
\label{fig:hyper}
\end{figure}

% one parameter to be considered: the area of receptor to get information
\
\newline
\indent In our proposed method, there are two key hyperparameters to choose: the size of our virtual map and the evaporation coefficient of pheromones. In Figure \ref{fig:hyper}(a), models are trained with virtual map sizes 6, 8 and 10 respectively. The result indicates that PooL is not sensitive to this parameter. Figure \ref{fig:hyper}(b) indicates that when the evaporation coefficient is between 0.1-0.9, the average reward after model convergence is almost the same. A value of 0.1 or 0.9 may slightly reduce the convergence speed of the model. Therefore, we recommend a value of about 0.5 according to the actual environment. If the evaporation coefficient is too large, the model can not effectively extract historical information. If the evaporation coefficient is too small, the information reflected by pheromone will lag.

\section{Conclusion}
% swarm intelligence or Collective intelligence
In this paper, we propose PooL, a pheromone-based indirect communication framework applied to MARL cooperation task. Inspired by swarm intelligence algorithms like ACO, PooL inherits the advantage of swarm intelligence algorithms that can be applied to large-scale agent coordination. Unlike ACO, pheromones released by agents are the output of DRL algorithms. Q-Learning is taken as an example to realize our framework, but other reinforcement learning algorithms can be applied to PooL in the same way. In our proposed framework, the information of agents in the environment is organized and dimensionally reduced by the pheromone update mechanism. Therefore, the cost of communication between agents is also very small. PooL is evaluated with different MARL algorithms in various environments proposed by Pettingzoo which contains hundreds of agents. Experimental results show that PooL converges faster and has higher rewards than other methods. Its superior performance with low cost in game environments shows that it has the potential to be applied in more complex real-world scenes with a bandwidth limit. PooL's success indicates that rule-based swarm intelligence control algorithms are promising to combine with learning-based models for large-scale agents' coordination.

\begin{acks}

This work was supported by NSF China under grand xxxxxxxx.
\end{acks}

%%%%%%%%%%%%%%%%%%%%%%%%%%%%%%%%%%%%%%%%%%%%%%%%%%%%%%%%%%%%%%%%%%%%%%%%

%%% The next two lines define, first, the bibliography style to be 
%%% applied, and, second, the bibliography file to be used.

\bibliographystyle{ACM-Reference-Format} 
\bibliography{sample}


\begin{thebibliography}{37}

%%% the Web address.

\ifx \showCODEN    \undefined \def \showCODEN     #1{\unskip}     \fi
\ifx \showDOI      \undefined \def \showDOI       #1{#1}\fi
\ifx \showISBNx    \undefined \def \showISBNx     #1{\unskip}     \fi
\ifx \showISBNxiii \undefined \def \showISBNxiii  #1{\unskip}     \fi
\ifx \showISSN     \undefined \def \showISSN      #1{\unskip}     \fi
\ifx \showLCCN     \undefined \def \showLCCN      #1{\unskip}     \fi
\ifx \shownote     \undefined \def \shownote      #1{#1}          \fi
\ifx \showarticletitle \undefined \def \showarticletitle #1{#1}   \fi
\ifx \showURL      \undefined \def \showURL       {\relax}        \fi
% invisible to TeX
\providecommand\bibfield[2]{#2}
\providecommand\bibinfo[2]{#2}
\providecommand\natexlab[1]{#1}
\providecommand\showeprint[2][]{arXiv:#2}

\bibitem[\protect\citeauthoryear{Bourenane, Mellouk, and
  Benhamamouch}{Bourenane et~al\mbox{.}}{2007}]%
        {bourenane2007reinforcement}
\bibfield{author}{\bibinfo{person}{Malika Bourenane},
  \bibinfo{person}{Abdelhamid Mellouk}, {and} \bibinfo{person}{Djilali
  Benhamamouch}.} \bibinfo{year}{2007}\natexlab{}.
\newblock \showarticletitle{Reinforcement learning in multi-agent environment
  and ant colony for packet scheduling in routers}. In
  \bibinfo{booktitle}{\emph{Proceedings of the 5th ACM international workshop
  on Mobility management and wireless access}}. \bibinfo{pages}{137--143}.
\newblock


\bibitem[\protect\citeauthoryear{Bu{\c{s}}oniu, Babu{\v{s}}ka, and
  De~Schutter}{Bu{\c{s}}oniu et~al\mbox{.}}{2010}]%
        {bucsoniu2010multi}
\bibfield{author}{\bibinfo{person}{Lucian Bu{\c{s}}oniu},
  \bibinfo{person}{Robert Babu{\v{s}}ka}, {and} \bibinfo{person}{Bart
  De~Schutter}.} \bibinfo{year}{2010}\natexlab{}.
\newblock \showarticletitle{Multi-agent reinforcement learning: An overview}.
\newblock \bibinfo{journal}{\emph{Innovations in multi-agent systems and
  applications-1}} (\bibinfo{year}{2010}), \bibinfo{pages}{183--221}.
\newblock


\bibitem[\protect\citeauthoryear{Dai, Long, Zhang, and Gong}{Dai
  et~al\mbox{.}}{2019}]%
        {dai2019mobile}
\bibfield{author}{\bibinfo{person}{Xiaolin Dai}, \bibinfo{person}{Shuai Long},
  \bibinfo{person}{Zhiwen Zhang}, {and} \bibinfo{person}{Dawei Gong}.}
  \bibinfo{year}{2019}\natexlab{}.
\newblock \showarticletitle{Mobile robot path planning based on ant colony
  algorithm with A* heuristic method}.
\newblock \bibinfo{journal}{\emph{Frontiers in neurorobotics}}
  \bibinfo{volume}{13} (\bibinfo{year}{2019}), \bibinfo{pages}{15}.
\newblock


\bibitem[\protect\citeauthoryear{Das, Gervet, Romoff, Batra, Parikh, Rabbat,
  and Pineau}{Das et~al\mbox{.}}{2019}]%
        {das2019tarmac}
\bibfield{author}{\bibinfo{person}{Abhishek Das},
  \bibinfo{person}{Th{\'e}ophile Gervet}, \bibinfo{person}{Joshua Romoff},
  \bibinfo{person}{Dhruv Batra}, \bibinfo{person}{Devi Parikh},
  \bibinfo{person}{Mike Rabbat}, {and} \bibinfo{person}{Joelle Pineau}.}
  \bibinfo{year}{2019}\natexlab{}.
\newblock \showarticletitle{Tarmac: Targeted multi-agent communication}. In
  \bibinfo{booktitle}{\emph{International Conference on Machine Learning}}.
  PMLR, \bibinfo{pages}{1538--1546}.
\newblock


\bibitem[\protect\citeauthoryear{Di~Caro and Dorigo}{Di~Caro and
  Dorigo}{1998}]%
        {di1998antnet}
\bibfield{author}{\bibinfo{person}{Gianni Di~Caro} {and} \bibinfo{person}{Marco
  Dorigo}.} \bibinfo{year}{1998}\natexlab{}.
\newblock \showarticletitle{AntNet: Distributed stigmergetic control for
  communications networks}.
\newblock \bibinfo{journal}{\emph{Journal of Artificial Intelligence Research}}
   \bibinfo{volume}{9} (\bibinfo{year}{1998}), \bibinfo{pages}{317--365}.
\newblock


\bibitem[\protect\citeauthoryear{Dorigo, Birattari, and Stutzle}{Dorigo
  et~al\mbox{.}}{2006}]%
        {dorigo2006ant}
\bibfield{author}{\bibinfo{person}{Marco Dorigo}, \bibinfo{person}{Mauro
  Birattari}, {and} \bibinfo{person}{Thomas Stutzle}.}
  \bibinfo{year}{2006}\natexlab{}.
\newblock \showarticletitle{Ant colony optimization}.
\newblock \bibinfo{journal}{\emph{IEEE computational intelligence magazine}}
  \bibinfo{volume}{1}, \bibinfo{number}{4} (\bibinfo{year}{2006}),
  \bibinfo{pages}{28--39}.
\newblock


\bibitem[\protect\citeauthoryear{Foerster, Farquhar, Afouras, Nardelli, and
  Whiteson}{Foerster et~al\mbox{.}}{2018}]%
        {foerster2018counterfactual}
\bibfield{author}{\bibinfo{person}{Jakob Foerster}, \bibinfo{person}{Gregory
  Farquhar}, \bibinfo{person}{Triantafyllos Afouras}, \bibinfo{person}{Nantas
  Nardelli}, {and} \bibinfo{person}{Shimon Whiteson}.}
  \bibinfo{year}{2018}\natexlab{}.
\newblock \showarticletitle{Counterfactual multi-agent policy gradients}. In
  \bibinfo{booktitle}{\emph{Proceedings of the AAAI Conference on Artificial
  Intelligence}}, Vol.~\bibinfo{volume}{32}.
\newblock


\bibitem[\protect\citeauthoryear{Haarnoja, Tang, Abbeel, and Levine}{Haarnoja
  et~al\mbox{.}}{2017}]%
        {haarnoja2017reinforcement}
\bibfield{author}{\bibinfo{person}{Tuomas Haarnoja}, \bibinfo{person}{Haoran
  Tang}, \bibinfo{person}{Pieter Abbeel}, {and} \bibinfo{person}{Sergey
  Levine}.} \bibinfo{year}{2017}\natexlab{}.
\newblock \showarticletitle{Reinforcement learning with deep energy-based
  policies}. In \bibinfo{booktitle}{\emph{International Conference on Machine
  Learning}}. PMLR, \bibinfo{pages}{1352--1361}.
\newblock


\bibitem[\protect\citeauthoryear{Hansen, Bernstein, and Zilberstein}{Hansen
  et~al\mbox{.}}{2004}]%
        {hansen2004dynamic}
\bibfield{author}{\bibinfo{person}{Eric~A Hansen}, \bibinfo{person}{Daniel~S
  Bernstein}, {and} \bibinfo{person}{Shlomo Zilberstein}.}
  \bibinfo{year}{2004}\natexlab{}.
\newblock \showarticletitle{Dynamic programming for partially observable
  stochastic games}. In \bibinfo{booktitle}{\emph{AAAI}},
  Vol.~\bibinfo{volume}{4}. \bibinfo{pages}{709--715}.
\newblock


\bibitem[\protect\citeauthoryear{Heylighen}{Heylighen}{2015}]%
        {heylighen2015stigmergy}
\bibfield{author}{\bibinfo{person}{Francis Heylighen}.}
  \bibinfo{year}{2015}\natexlab{}.
\newblock \showarticletitle{Stigmergy as a Universal Coordination Mechanism:
  components, varieties and applications}.
\newblock \bibinfo{journal}{\emph{Human Stigmergy: Theoretical Developments and
  New Applications; Springer: New York, NY, USA}} (\bibinfo{year}{2015}).
\newblock


\bibitem[\protect\citeauthoryear{Hoshen}{Hoshen}{2018}]%
        {hoshen2018vain}
\bibfield{author}{\bibinfo{person}{Yedid Hoshen}.}
  \bibinfo{year}{2018}\natexlab{}.
\newblock \showarticletitle{VAIN: Attentional Multi-agent Predictive Modeling}.
\newblock \bibinfo{journal}{\emph{ADVANCES IN NEURAL INFORMATION PROCESSING
  SYSTEMS 30 (NIPS 2017)}} (\bibinfo{year}{2018}), \bibinfo{pages}{2701--2711}.
\newblock


\bibitem[\protect\citeauthoryear{Jiang, Dun, Huang, and Lu}{Jiang
  et~al\mbox{.}}{2020}]%
        {jiang2020graph}
\bibfield{author}{\bibinfo{person}{Jiechuan Jiang}, \bibinfo{person}{Chen Dun},
  \bibinfo{person}{Tiejun Huang}, {and} \bibinfo{person}{Zongqing Lu}.}
  \bibinfo{year}{2020}\natexlab{}.
\newblock \showarticletitle{Graph Convolutional Reinforcement Learning}.
\newblock \bibinfo{journal}{\emph{ICLR}} (\bibinfo{year}{2020}).
\newblock


\bibitem[\protect\citeauthoryear{Jiang and Lu}{Jiang and Lu}{2018}]%
        {jiang2018learning}
\bibfield{author}{\bibinfo{person}{Jiechuan Jiang} {and}
  \bibinfo{person}{Zongqing Lu}.} \bibinfo{year}{2018}\natexlab{}.
\newblock \showarticletitle{Learning Attentional Communication for Multi-Agent
  Cooperation}.
\newblock \bibinfo{journal}{\emph{ADVANCES IN NEURAL INFORMATION PROCESSING
  SYSTEMS 31 (NIPS 2018)}} (\bibinfo{year}{2018}), \bibinfo{pages}{7254--7264}.
\newblock


\bibitem[\protect\citeauthoryear{Kennedy}{Kennedy}{2006}]%
        {kennedy2006swarm}
\bibfield{author}{\bibinfo{person}{James Kennedy}.}
  \bibinfo{year}{2006}\natexlab{}.
\newblock \showarticletitle{Swarm intelligence}.
\newblock In \bibinfo{booktitle}{\emph{Handbook of nature-inspired and
  innovative computing}}. \bibinfo{publisher}{Springer},
  \bibinfo{pages}{187--219}.
\newblock


\bibitem[\protect\citeauthoryear{Konatowski and Paw{\l}owski}{Konatowski and
  Paw{\l}owski}{2018}]%
        {konatowski2018ant}
\bibfield{author}{\bibinfo{person}{Stanis{\l}aw Konatowski} {and}
  \bibinfo{person}{Piotr Paw{\l}owski}.} \bibinfo{year}{2018}\natexlab{}.
\newblock \showarticletitle{Ant colony optimization algorithm for UAV path
  planning}. In \bibinfo{booktitle}{\emph{2018 14th International Conference on
  Advanced Trends in Radioelecrtronics, Telecommunications and Computer
  Engineering (TCSET)}}. IEEE, \bibinfo{pages}{177--182}.
\newblock


\bibitem[\protect\citeauthoryear{lillicrap, hunt, pritzel, heess, erez, tassa,
  silver, and wierstra}{lillicrap et~al\mbox{.}}{2016}]%
        {lillicrap2016continuous}
\bibfield{author}{\bibinfo{person}{p~timothy lillicrap},
  \bibinfo{person}{j~jonathan hunt}, \bibinfo{person}{alexander pritzel},
  \bibinfo{person}{nicolas heess}, \bibinfo{person}{tom erez},
  \bibinfo{person}{yuval tassa}, \bibinfo{person}{david silver}, {and}
  \bibinfo{person}{daan wierstra}.} \bibinfo{year}{2016}\natexlab{}.
\newblock \showarticletitle{Continuous control with deep reinforcement
  learning}.
\newblock \bibinfo{journal}{\emph{CoRR}} (\bibinfo{year}{2016}).
\newblock


\bibitem[\protect\citeauthoryear{Lowe, Wu, Tamar, Harb, Abbeel, and
  Mordatch}{Lowe et~al\mbox{.}}{2020}]%
        {lowe2020multi-agent}
\bibfield{author}{\bibinfo{person}{Ryan Lowe}, \bibinfo{person}{Yi Wu},
  \bibinfo{person}{Aviv Tamar}, \bibinfo{person}{Jean Harb},
  \bibinfo{person}{Pieter Abbeel}, {and} \bibinfo{person}{Igor Mordatch}.}
  \bibinfo{year}{2020}\natexlab{}.
\newblock \showarticletitle{Multi-Agent Actor-Critic for Mixed
  Cooperative-Competitive Environments}.
\newblock \bibinfo{journal}{\emph{ADVANCES IN NEURAL INFORMATION PROCESSING
  SYSTEMS 30 (NIPS 2017)}} (\bibinfo{year}{2020}), \bibinfo{pages}{6379--6390}.
\newblock


\bibitem[\protect\citeauthoryear{Luo*, Yang*, Li, Zhou, Zhang, and Wang}{Luo*
  et~al\mbox{.}}{2018}]%
        {luo*2018mean}
\bibfield{author}{\bibinfo{person}{Rui Luo*}, \bibinfo{person}{Yaodong Yang*},
  \bibinfo{person}{Minne Li}, \bibinfo{person}{Ming Zhou},
  \bibinfo{person}{Weinan Zhang}, {and} \bibinfo{person}{Jun Wang}.}
  \bibinfo{year}{2018}\natexlab{}.
\newblock \showarticletitle{Mean Field Multi Agent Reinforcement Learning}.
\newblock \bibinfo{journal}{\emph{The 35th International Conference on Machine
  Learning (ICML'18), PMLR}} (\bibinfo{year}{2018}),
  \bibinfo{pages}{5567--5576}.
\newblock


\bibitem[\protect\citeauthoryear{Maru, Enoki, Nakao, Yamamoto, Yamaguchi, and
  Oguchi}{Maru et~al\mbox{.}}{2016}]%
        {maru2016qoe}
\bibfield{author}{\bibinfo{person}{Chihiro Maru}, \bibinfo{person}{Miki Enoki},
  \bibinfo{person}{Akihiro Nakao}, \bibinfo{person}{Shu Yamamoto},
  \bibinfo{person}{Saneyasu Yamaguchi}, {and} \bibinfo{person}{Masato Oguchi}.}
  \bibinfo{year}{2016}\natexlab{}.
\newblock \showarticletitle{QoE Control of Network Using Collective
  Intelligence of SNS in Large-Scale Disasters}. In
  \bibinfo{booktitle}{\emph{2016 IEEE International Conference on Computer and
  Information Technology (CIT)}}. IEEE, \bibinfo{pages}{57--64}.
\newblock


\bibitem[\protect\citeauthoryear{Matta, Cardarilli, Di~Nunzio, Fazzolari,
  Giardino, Re, Silvestri, and Span{\`o}}{Matta et~al\mbox{.}}{2019}]%
        {matta2019q}
\bibfield{author}{\bibinfo{person}{Marco Matta}, \bibinfo{person}{Gian~Carlo
  Cardarilli}, \bibinfo{person}{Luca Di~Nunzio}, \bibinfo{person}{Rocco
  Fazzolari}, \bibinfo{person}{Daniele Giardino}, \bibinfo{person}{M Re},
  \bibinfo{person}{F Silvestri}, {and} \bibinfo{person}{S Span{\`o}}.}
  \bibinfo{year}{2019}\natexlab{}.
\newblock \showarticletitle{Q-RTS: a real-time swarm intelligence based on
  multi-agent Q-learning}.
\newblock \bibinfo{journal}{\emph{Electronics Letters}} \bibinfo{volume}{55},
  \bibinfo{number}{10} (\bibinfo{year}{2019}), \bibinfo{pages}{589--591}.
\newblock


\bibitem[\protect\citeauthoryear{Meerza, Islam, and Uzzal}{Meerza
  et~al\mbox{.}}{2019}]%
        {meerza2019q}
\bibfield{author}{\bibinfo{person}{Syed Irfan~Ali Meerza},
  \bibinfo{person}{Moinul Islam}, {and} \bibinfo{person}{Md~Mohiuddin Uzzal}.}
  \bibinfo{year}{2019}\natexlab{}.
\newblock \showarticletitle{Q-Learning Based Particle Swarm Optimization
  Algorithm for Optimal Path Planning of Swarm of Mobile Robots}. In
  \bibinfo{booktitle}{\emph{2019 1st International Conference on Advances in
  Science, Engineering and Robotics Technology (ICASERT)}}. IEEE,
  \bibinfo{pages}{1--5}.
\newblock


\bibitem[\protect\citeauthoryear{Mnih, Kavukcuoglu, Silver, Graves, Antonoglou,
  Wierstra, and Riedmiller}{Mnih et~al\mbox{.}}{2013}]%
        {mnih2013playing}
\bibfield{author}{\bibinfo{person}{Volodymyr Mnih}, \bibinfo{person}{Koray
  Kavukcuoglu}, \bibinfo{person}{David Silver}, \bibinfo{person}{Alex Graves},
  \bibinfo{person}{Ioannis Antonoglou}, \bibinfo{person}{Daan Wierstra}, {and}
  \bibinfo{person}{Martin Riedmiller}.} \bibinfo{year}{2013}\natexlab{}.
\newblock \showarticletitle{Playing Atari with Deep Reinforcement Learning}.
\newblock \bibinfo{journal}{\emph{CoRR}} (\bibinfo{year}{2013}).
\newblock


\bibitem[\protect\citeauthoryear{Mnih, Kavukcuoglu, Silver, Rusu, Veness,
  Bellemare, Graves, Riedmiller, Fidjeland, Ostrovski, et~al\mbox{.}}{Mnih
  et~al\mbox{.}}{2015}]%
        {mnih2015human}
\bibfield{author}{\bibinfo{person}{Volodymyr Mnih}, \bibinfo{person}{Koray
  Kavukcuoglu}, \bibinfo{person}{David Silver}, \bibinfo{person}{Andrei~A
  Rusu}, \bibinfo{person}{Joel Veness}, \bibinfo{person}{Marc~G Bellemare},
  \bibinfo{person}{Alex Graves}, \bibinfo{person}{Martin Riedmiller},
  \bibinfo{person}{Andreas~K Fidjeland}, \bibinfo{person}{Georg Ostrovski},
  {et~al\mbox{.}}} \bibinfo{year}{2015}\natexlab{}.
\newblock \showarticletitle{Human-level control through deep reinforcement
  learning}.
\newblock \bibinfo{journal}{\emph{nature}} \bibinfo{volume}{518},
  \bibinfo{number}{7540} (\bibinfo{year}{2015}), \bibinfo{pages}{529--533}.
\newblock


\bibitem[\protect\citeauthoryear{Monekosso and Remagnino}{Monekosso and
  Remagnino}{2001}]%
        {monekosso2001phe}
\bibfield{author}{\bibinfo{person}{Ndedi Monekosso} {and}
  \bibinfo{person}{Paolo Remagnino}.} \bibinfo{year}{2001}\natexlab{}.
\newblock \showarticletitle{Phe-Q: A pheromone based Q-learning}. In
  \bibinfo{booktitle}{\emph{Australian Joint Conference on Artificial
  Intelligence}}. Springer, \bibinfo{pages}{345--355}.
\newblock


\bibitem[\protect\citeauthoryear{Oliehoek, Spaan, and Vlassis}{Oliehoek
  et~al\mbox{.}}{2008}]%
        {oliehoek2008optimal}
\bibfield{author}{\bibinfo{person}{Frans~A Oliehoek},
  \bibinfo{person}{Matthijs~TJ Spaan}, {and} \bibinfo{person}{Nikos Vlassis}.}
  \bibinfo{year}{2008}\natexlab{}.
\newblock \showarticletitle{Optimal and approximate Q-value functions for
  decentralized POMDPs}.
\newblock \bibinfo{journal}{\emph{Journal of Artificial Intelligence Research}}
   \bibinfo{volume}{32} (\bibinfo{year}{2008}), \bibinfo{pages}{289--353}.
\newblock


\bibitem[\protect\citeauthoryear{Silver, Huang, Maddison, Guez, Sifre, Van
  Den~Driessche, Schrittwieser, Antonoglou, Panneershelvam, Lanctot,
  et~al\mbox{.}}{Silver et~al\mbox{.}}{2016}]%
        {silver2016mastering}
\bibfield{author}{\bibinfo{person}{David Silver}, \bibinfo{person}{Aja Huang},
  \bibinfo{person}{Chris~J Maddison}, \bibinfo{person}{Arthur Guez},
  \bibinfo{person}{Laurent Sifre}, \bibinfo{person}{George Van Den~Driessche},
  \bibinfo{person}{Julian Schrittwieser}, \bibinfo{person}{Ioannis Antonoglou},
  \bibinfo{person}{Veda Panneershelvam}, \bibinfo{person}{Marc Lanctot},
  {et~al\mbox{.}}} \bibinfo{year}{2016}\natexlab{}.
\newblock \showarticletitle{Mastering the game of Go with deep neural networks
  and tree search}.
\newblock \bibinfo{journal}{\emph{nature}} \bibinfo{volume}{529},
  \bibinfo{number}{7587} (\bibinfo{year}{2016}), \bibinfo{pages}{484--489}.
\newblock


\bibitem[\protect\citeauthoryear{Sukhbaatar, Szlam, and Fergus}{Sukhbaatar
  et~al\mbox{.}}{2016}]%
        {sukhbaatar2016learning}
\bibfield{author}{\bibinfo{person}{Sainbayar Sukhbaatar},
  \bibinfo{person}{Arthur Szlam}, {and} \bibinfo{person}{Rob Fergus}.}
  \bibinfo{year}{2016}\natexlab{}.
\newblock \showarticletitle{Learning Multiagent Communication with
  Backpropagation}.
\newblock \bibinfo{journal}{\emph{ADVANCES IN NEURAL INFORMATION PROCESSING
  SYSTEMS 29 (NIPS 2016)}} (\bibinfo{year}{2016}), \bibinfo{pages}{2252--2260}.
\newblock


\bibitem[\protect\citeauthoryear{Terry, Black, Grammel, Jayakumar, Hari,
  Sulivan, Santos, Perez, Horsch, Dieffendahl, Williams, Lokesh, Sullivan, and
  Ravi}{Terry et~al\mbox{.}}{2020}]%
        {terry2020pettingzoo}
\bibfield{author}{\bibinfo{person}{J.~K Terry}, \bibinfo{person}{Benjamin
  Black}, \bibinfo{person}{Nathaniel Grammel}, \bibinfo{person}{Mario
  Jayakumar}, \bibinfo{person}{Ananth Hari}, \bibinfo{person}{Ryan Sulivan},
  \bibinfo{person}{Luis Santos}, \bibinfo{person}{Rodrigo Perez},
  \bibinfo{person}{Caroline Horsch}, \bibinfo{person}{Clemens Dieffendahl},
  \bibinfo{person}{Niall~L Williams}, \bibinfo{person}{Yashas Lokesh},
  \bibinfo{person}{Ryan Sullivan}, {and} \bibinfo{person}{Praveen Ravi}.}
  \bibinfo{year}{2020}\natexlab{}.
\newblock \showarticletitle{PettingZoo: Gym for Multi-Agent Reinforcement
  Learning}.
\newblock \bibinfo{journal}{\emph{arXiv preprint arXiv:2009.14471}}
  (\bibinfo{year}{2020}).
\newblock


\bibitem[\protect\citeauthoryear{Wang, Yang, Liu, Hao, Hao, Hu, Chen, Fan, and
  Gao}{Wang et~al\mbox{.}}{2020}]%
        {wang2020few}
\bibfield{author}{\bibinfo{person}{Weixun Wang}, \bibinfo{person}{Tianpei
  Yang}, \bibinfo{person}{Yong Liu}, \bibinfo{person}{Jianye Hao},
  \bibinfo{person}{Xiaotian Hao}, \bibinfo{person}{Yujing Hu},
  \bibinfo{person}{Yingfeng Chen}, \bibinfo{person}{Changjie Fan}, {and}
  \bibinfo{person}{Yang Gao}.} \bibinfo{year}{2020}\natexlab{}.
\newblock \showarticletitle{From few to more: Large-scale dynamic multiagent
  curriculum learning}. In \bibinfo{booktitle}{\emph{Proceedings of the AAAI
  Conference on Artificial Intelligence}}, Vol.~\bibinfo{volume}{34}.
  \bibinfo{pages}{7293--7300}.
\newblock


\bibitem[\protect\citeauthoryear{Wei, Wicke, Freelan, and Luke}{Wei
  et~al\mbox{.}}{2018}]%
        {wei2018multiagent}
\bibfield{author}{\bibinfo{person}{Ermo Wei}, \bibinfo{person}{Drew Wicke},
  \bibinfo{person}{David Freelan}, {and} \bibinfo{person}{Sean Luke}.}
  \bibinfo{year}{2018}\natexlab{}.
\newblock \showarticletitle{Multiagent soft q-learning}. In
  \bibinfo{booktitle}{\emph{2018 AAAI Spring Symposium Series}}.
\newblock


\bibitem[\protect\citeauthoryear{Xu, Li, Zhao, and Zhang}{Xu
  et~al\mbox{.}}{2021}]%
        {xu2021stigmergic}
\bibfield{author}{\bibinfo{person}{Xing Xu}, \bibinfo{person}{Rongpeng Li},
  \bibinfo{person}{Zhifeng Zhao}, {and} \bibinfo{person}{Honggang Zhang}.}
  \bibinfo{year}{2021}\natexlab{}.
\newblock \showarticletitle{Stigmergic Independent Reinforcement Learning for
  Multiagent Collaboration}.
\newblock \bibinfo{journal}{\emph{IEEE Transactions on Neural Networks and
  Learning Systems}} (\bibinfo{year}{2021}).
\newblock


\bibitem[\protect\citeauthoryear{Xu, Zhao, Li, and Zhang}{Xu
  et~al\mbox{.}}{2019}]%
        {xu2019brain}
\bibfield{author}{\bibinfo{person}{Xing Xu}, \bibinfo{person}{Zhifeng Zhao},
  \bibinfo{person}{Rongpeng Li}, {and} \bibinfo{person}{Honggang Zhang}.}
  \bibinfo{year}{2019}\natexlab{}.
\newblock \showarticletitle{Brain-inspired stigmergy learning}.
\newblock \bibinfo{journal}{\emph{IEEE Access}}  \bibinfo{volume}{7}
  (\bibinfo{year}{2019}), \bibinfo{pages}{54410--54424}.
\newblock


\bibitem[\protect\citeauthoryear{Yang, Luo, Li, Zhou, Zhang, and Wang}{Yang
  et~al\mbox{.}}{2018}]%
        {yang2018mean}
\bibfield{author}{\bibinfo{person}{Yaodong Yang}, \bibinfo{person}{Rui Luo},
  \bibinfo{person}{Minne Li}, \bibinfo{person}{Ming Zhou},
  \bibinfo{person}{Weinan Zhang}, {and} \bibinfo{person}{Jun Wang}.}
  \bibinfo{year}{2018}\natexlab{}.
\newblock \showarticletitle{Mean field multi-agent reinforcement learning}. In
  \bibinfo{booktitle}{\emph{International Conference on Machine Learning}}.
  PMLR, \bibinfo{pages}{5571--5580}.
\newblock


\bibitem[\protect\citeauthoryear{Zhao and Ma}{Zhao and Ma}{2019}]%
        {zhao2019learning}
\bibfield{author}{\bibinfo{person}{Yuhang Zhao} {and} \bibinfo{person}{Xiujun
  Ma}.} \bibinfo{year}{2019}\natexlab{}.
\newblock \showarticletitle{Learning Efficient Communication in Cooperative
  Multi-Agent Environment.}. In \bibinfo{booktitle}{\emph{AAMAS}}.
  \bibinfo{pages}{2321--2323}.
\newblock


\bibitem[\protect\citeauthoryear{Zheng, Yang, Cai, Zhou, Zhang, Wang, and
  Yu}{Zheng et~al\mbox{.}}{2018}]%
        {zheng2018magent}
\bibfield{author}{\bibinfo{person}{Lianmin Zheng}, \bibinfo{person}{Jiacheng
  Yang}, \bibinfo{person}{Han Cai}, \bibinfo{person}{Ming Zhou},
  \bibinfo{person}{Weinan Zhang}, \bibinfo{person}{Jun Wang}, {and}
  \bibinfo{person}{Yong Yu}.} \bibinfo{year}{2018}\natexlab{}.
\newblock \showarticletitle{Magent: A many-agent reinforcement learning
  platform for artificial collective intelligence}. In
  \bibinfo{booktitle}{\emph{Proceedings of the AAAI Conference on Artificial
  Intelligence}}, Vol.~\bibinfo{volume}{32}.
\newblock


\bibitem[\protect\citeauthoryear{Zhou, Cui, Hu, Zhang, Yang, Liu, Wang, Li, and
  Sun}{Zhou et~al\mbox{.}}{2020}]%
        {zhou2020graph}
\bibfield{author}{\bibinfo{person}{Jie Zhou}, \bibinfo{person}{Ganqu Cui},
  \bibinfo{person}{Shengding Hu}, \bibinfo{person}{Zhengyan Zhang},
  \bibinfo{person}{Cheng Yang}, \bibinfo{person}{Zhiyuan Liu},
  \bibinfo{person}{Lifeng Wang}, \bibinfo{person}{Changcheng Li}, {and}
  \bibinfo{person}{Maosong Sun}.} \bibinfo{year}{2020}\natexlab{}.
\newblock \showarticletitle{Graph neural networks: A review of methods and
  applications}.
\newblock \bibinfo{journal}{\emph{AI Open}}  \bibinfo{volume}{1}
  (\bibinfo{year}{2020}), \bibinfo{pages}{57--81}.
\newblock


\bibitem[\protect\citeauthoryear{Zhou, Chen, Wen, Yang, Su, Zhang, Zhang, and
  Wang}{Zhou et~al\mbox{.}}{2019}]%
        {zhou2019factorized}
\bibfield{author}{\bibinfo{person}{Ming Zhou}, \bibinfo{person}{Yong Chen},
  \bibinfo{person}{Ying Wen}, \bibinfo{person}{Yaodong Yang},
  \bibinfo{person}{Yufeng Su}, \bibinfo{person}{Weinan Zhang},
  \bibinfo{person}{Dell Zhang}, {and} \bibinfo{person}{Jun Wang}.}
  \bibinfo{year}{2019}\natexlab{}.
\newblock \showarticletitle{Factorized q-learning for large-scale multi-agent
  systems}. In \bibinfo{booktitle}{\emph{Proceedings of the First International
  Conference on Distributed Artificial Intelligence}}. \bibinfo{pages}{1--7}.
\newblock


\end{thebibliography}

%%%%%%%%%%%%%%%%%%%%%%%%%%%%%%%%%%%%%%%%%%%%%%%%%%%%%%%%%%%%%%%%%%%%%%%%

\end{document}